%% file: collas2025_conference.tex
\documentclass{article} 

\usepackage{ifthen}
\newboolean{public_version}
\setboolean{public_version}{true}  

\usepackage{collas2025_conference,times} 
\usepackage{easyReview}

\usepackage{booktabs} 
\usepackage{comment}

\usepackage{appendix}
\usepackage{enumitem}
\usepackage{subcaption}
\usepackage{pifont}
\usepackage{wrapfig}

\input{math_commands.tex}

\usepackage{hyperref}
\hypersetup{
    colorlinks=true,
    linkcolor=red,
    filecolor=magenta,
    urlcolor=blue,
    citecolor=purple,
    pdftitle={Overleaf Example},
    pdfpagemode=FullScreen,
    }

\title{A Good Start Matters: Enhancing Continual Learning with Data-Driven Weight Initialization}

\author{Md Yousuf Harun\\
Rochester Institute of Technology\\
United States of America\\
\texttt{mh1023@rit.edu} \\
\And 
Christopher Kanan \\
University of Rochester \\
United States of America\\
\texttt{ckanan@cs.rochester.edu}\\
}

\collasfinalcopy 

\begin{document}

\maketitle



\input{sections/0-abstract}
\input{sections/1-introduction}
\input{sections/2-background}

\input{sections/3-method}
\input{sections/4-experiments}
\input{sections/5-discussion}

\input{sections/6-conclusion}

\ifthenelse{\boolean{public_version}}{
\subsubsection*{Acknowledgments} This work was partly supported by NSF awards \#2326491, \#2125362, and \#2317706. The views and conclusions contained herein are those of the authors and should not be interpreted as representing any sponsor's official policies or endorsements.

}

{
    \small
    \bibliographystyle{collas2025_conference}
    \bibliography{collas2025_conference}
}

\input{sections/supplemental}





\end{document}

%% file: math_commands.tex
\usepackage{amsmath,amsfonts,bm}

\def\eqref#1{equation~\ref{#1}}

\def\1{\bm{1}}

\DeclareMathAlphabet{\mathsfit}{\encodingdefault}{\sfdefault}{m}{sl}
\SetMathAlphabet{\mathsfit}{bold}{\encodingdefault}{\sfdefault}{bx}{n}



%% file: sections/0-abstract.tex
\begin{abstract}
To adapt to real-world data streams, continual learning (CL) systems must rapidly learn new concepts while preserving and utilizing prior knowledge. 
When it comes to adding new information to continually-trained deep neural networks (DNNs), 
classifier weights for newly encountered categories are typically initialized randomly, leading to high initial training loss (spikes) and instability. Consequently, achieving optimal convergence and accuracy requires prolonged training, increasing computational costs.
Inspired by Neural Collapse (NC), we propose a weight initialization strategy to improve learning efficiency in CL. In DNNs trained with mean-squared-error, NC gives rise to a Least-Square (LS) classifier in the last layer, whose weights can be analytically derived from learned features.
We leverage this LS formulation to initialize classifier weights in a data-driven manner, aligning them with the feature distribution rather than using random initialization. Our method mitigates initial loss spikes and accelerates adaptation to new tasks.
We evaluate our approach in large-scale CL settings, demonstrating faster adaptation and improved CL performance.
\end{abstract}

%% file: sections/1-introduction.tex
\section{Introduction}
\label{sec:intro}


Deep learning models excel in static environments where the data follows an independent and identically distributed (IID) assumption. However, in real-world scenarios, data distributions shift over time (non-IID), and new data arrives sequentially. Conventional deep neural networks (DNNs) struggle under such conditions, often requiring periodic re-training from scratch, which is not only computationally expensive but also contributes significantly to the carbon footprint of AI~\citep{schwartz2020green}. Despite frequent retraining from scratch, real-world models still suffer up to 40\% accuracy drops~\citep{mallick2022matchmaker}. 
Continual learning (CL) aims to address this inefficiency by enabling models to learn from evolving data streams while preserving previously acquired knowledge~\citep{parisi2019continual}.
CL is a promising solution to \textit{model decay}, where predictive performance deteriorates over time due to \textit{concept drift}—a shift in the meaning or distribution of target variables~\citep{tsymbal2004problem, gama2014survey,lu2018learning}.  
While much of the CL research community has focused on mitigating \textit{catastrophic forgetting}~\citep{mccloskey1989catastrophic}, there is growing interest in optimizing CL for \textit{computational efficiency}~\citep{ghunaim2023real,harun2023efficient,harun2023siesta, prabhu2023computationally, harun2024overcoming, 
harun2024grasp, verwimp2023continual}.

The increasing prevalence of large foundation models has shifted CL paradigm toward leveraging these models e.g., ImageNet-1K or ImageNet-21K pre-trained backbones~\citep{wang2022learning, wang2022dualprompt, smith2023coda, gao2023lae, mcdonnell2024ranpac, harun2024overcoming}. The challenge now lies in integrating new knowledge efficiently while maintaining and refining prior knowledge. However, prior studies~\citep{wang2022dualprompt, mirzadeh2022architecture, harun2024overcoming} reveal that naive use of pre-trained models does not inherently improve CL performance, and effectively adapting pre-trained models for CL remains an open challenge.  
A key research question in this adaptation process is: 
\textbf{\textit{how do we initialize the last-layer classifier weights for newly introduced categories?}}
Standard practice initializes new class weights \textit{randomly}, leading to high initial training loss, unstable gradient updates, and degraded performance~\citep{harun2024overcoming}.
Proper weight initialization is crucial for accelerating CL and enhancing plasticity~\citep{lyle2023understanding}.
As illustrated in Fig.~\ref{fig:vis_abs}, 
random weight initialization induces loss spikes when each new task is introduced, resulting in degraded performance and prolonged convergence time. 
This instability is especially problematic in real-world deployments where rapid model adaptation is critical. Despite its importance, weight initialization for new concepts remains an under-explored area in CL.

To address this challenge, we investigate \textit{data-driven weight initialization strategies based on feature statistics}. Inspired by the \textit{Neural Collapse (NC) phenomenon}, where deep networks naturally align last-layer weights with class means, we introduce a \textit{Least-Square (LS)}-based weight initialization that optimally sets new classifier weights using feature statistics.
LS weight initialization offers a principled alternative to random initialization, as it can be computed analytically \textit{solely} from penultimate-layer features, without requiring additional training or hyperparameters. 
Unlike random initialization, data-driven initialization provides a low-loss starting point (see Fig.~\ref{fig:vis_abs}), mitigating abrupt parameter shifts and stabilizing CL adaptation.
Our LS-based approach builds on \citet{han2022neural}, who derived LS weight formulation by minimizing the mean-squared-error (MSE) loss with weight decay. Their findings show that optimal classifier weights can be determined entirely from penultimate-layer features. By integrating LS-based initialization into CL frameworks, we demonstrate that it significantly reduces training loss spikes, leading to improved learning efficiency and performance.
\textbf{Our main contributions are summarized as follows:}
\begin{enumerate}[leftmargin=*] 
    \item We study the impact of weight initializations and training objectives on learning efficiency and CL performance.

    \item We propose a least-square-based weight initialization that optimally aligns classifier weights for newly encountered categories with their feature distributions.

    \item We empirically validate our data-driven initialization in large-scale CL settings, showing that our approach mitigates loss spikes and enhances adaptation efficiency.
    Additionally, we show that least-square initialization improves CL performance when integrated with various CL methods e.g., experience replay, EWC, and DER++.
    
\end{enumerate}


\input{figures/vis_abstract}

%% file: figures/vis_abstract.tex
\begin{figure}[t]
  \centering

\begin{subfigure}[b]{0.33\textwidth}
         \centering
         \includegraphics[width=\textwidth]{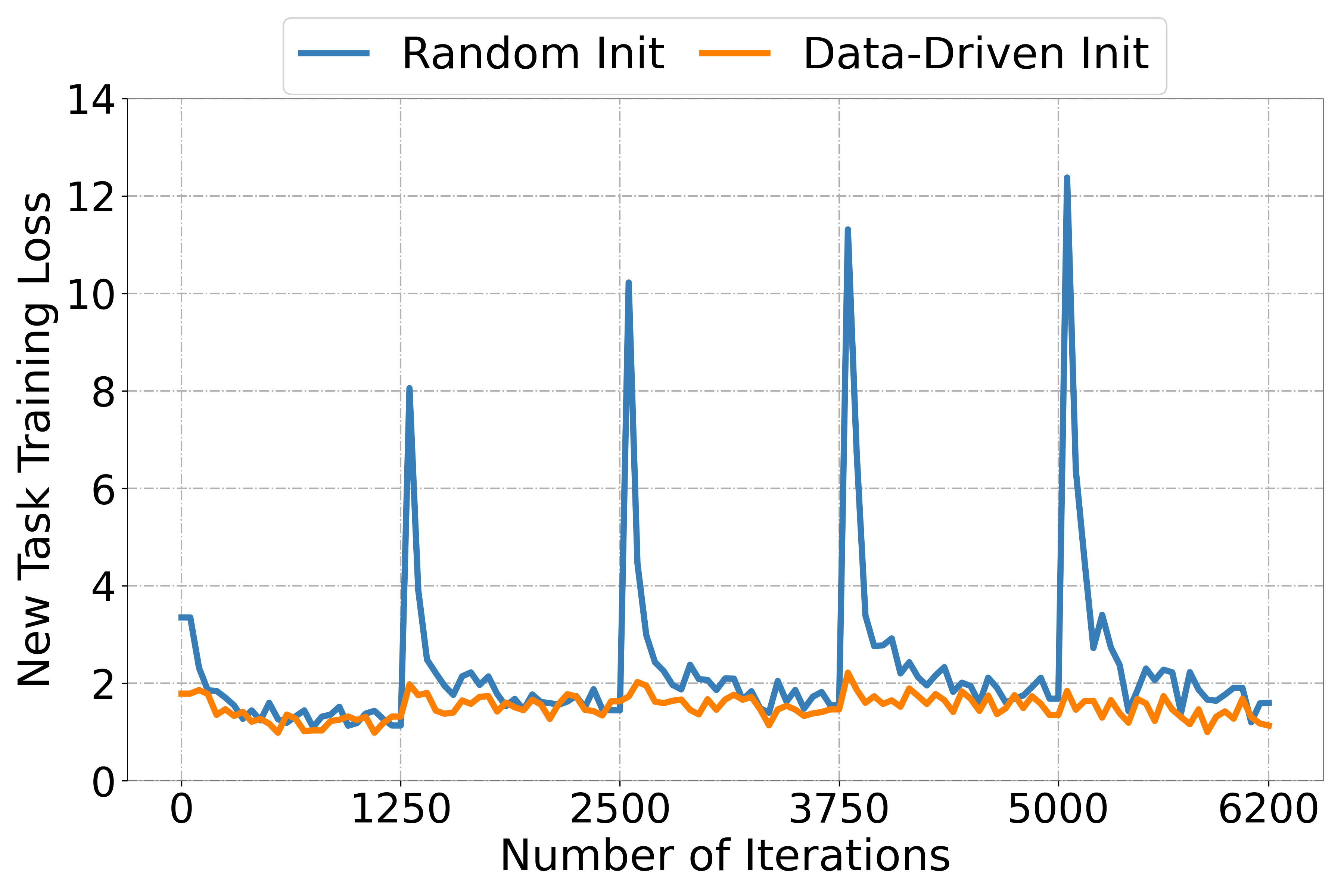}
         \caption{Training loss across CL tasks}
         \label{fig:train_loss}
     \end{subfigure}
     \hfill
      \begin{subfigure}[b]{0.33\textwidth}
         \centering
         \includegraphics[width=\textwidth]{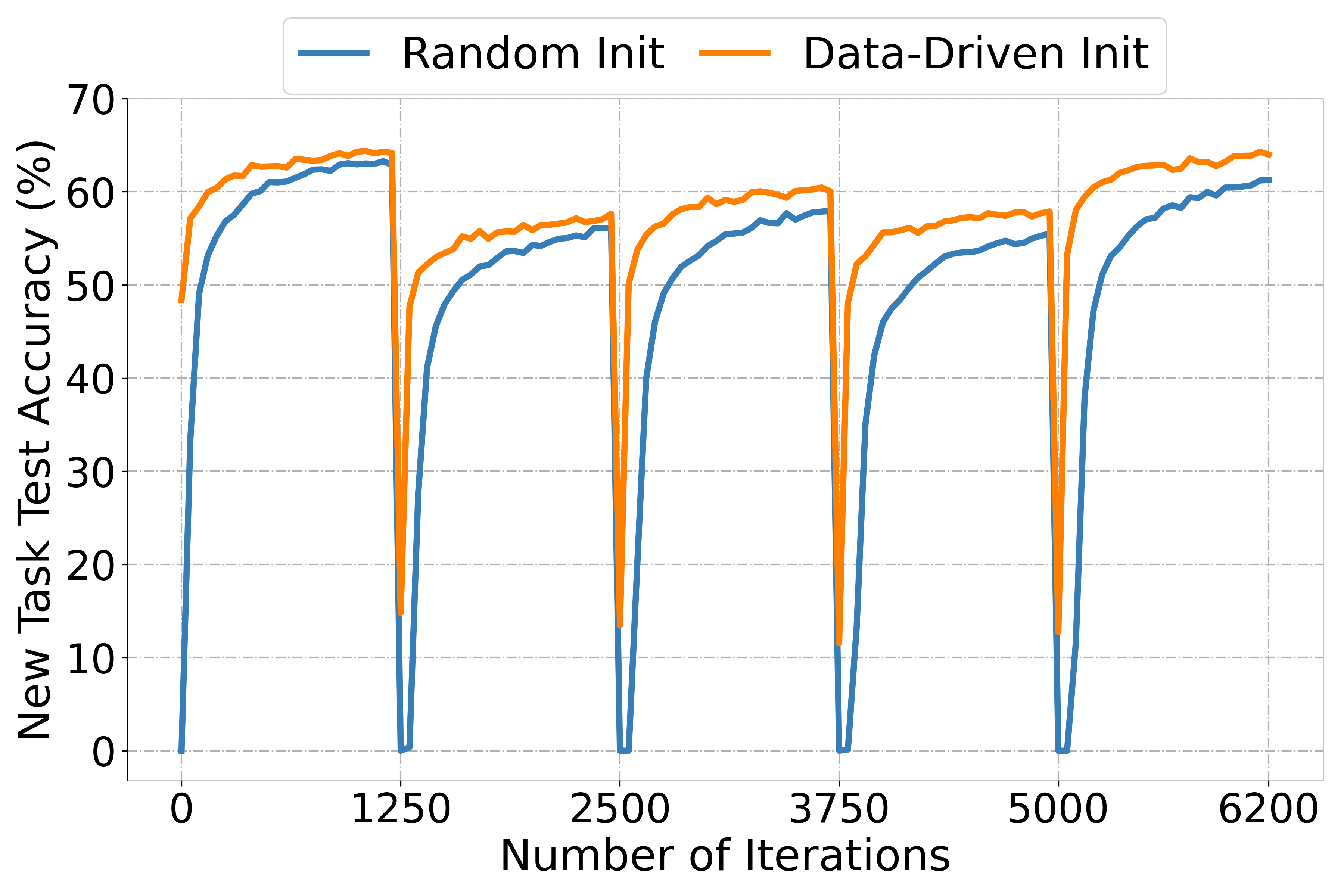}
         \caption{Test accuracy across CL tasks}
         \label{fig:test_acc}
     \end{subfigure}
     \hfill
      \begin{subfigure}[b]{0.33\textwidth}
         \centering
         \includegraphics[width=\textwidth]{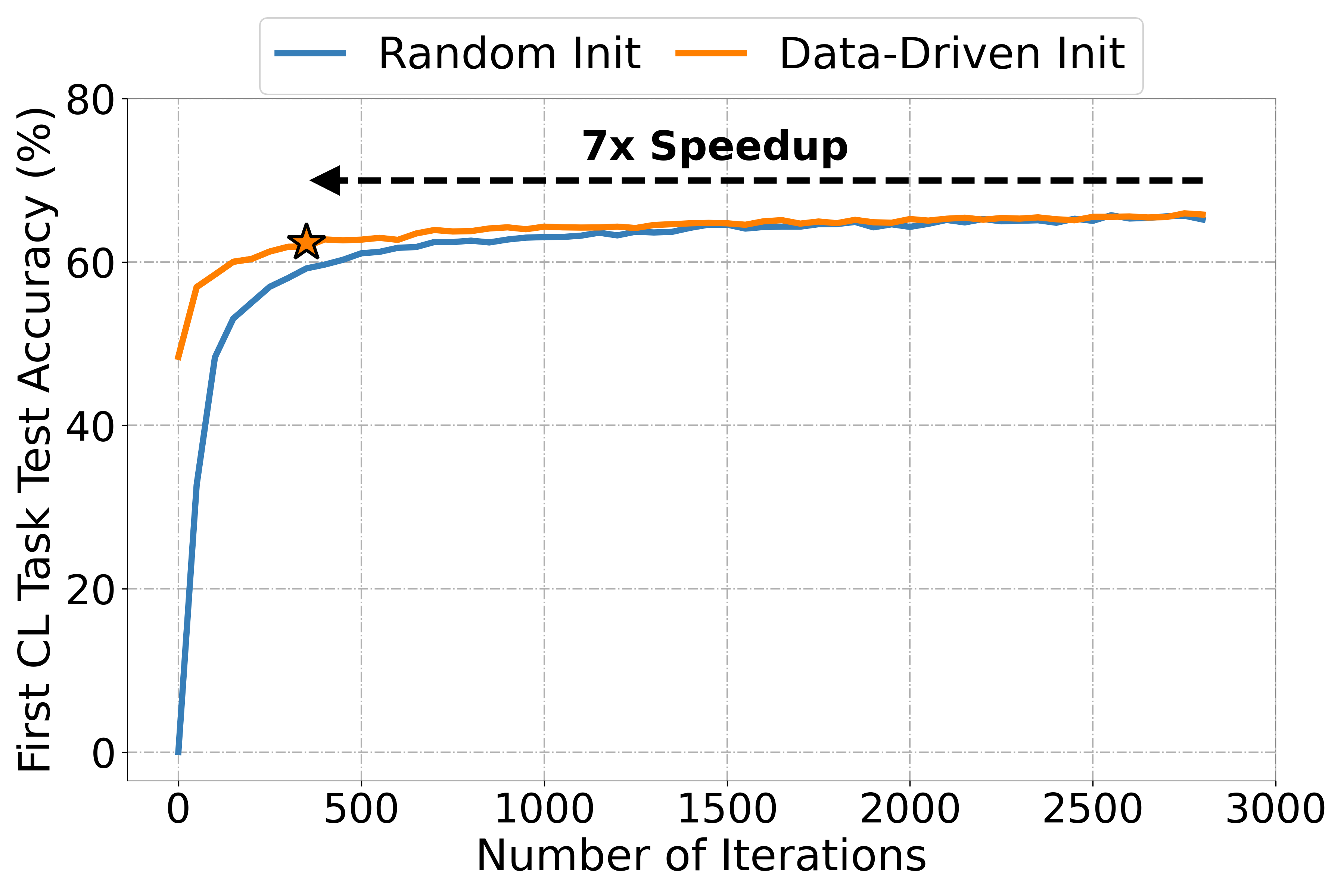}
         \caption{Efficiency in the first CL task}
         \label{fig:eff_gain}
     \end{subfigure}
   \caption{
   Data-driven initialization (least-square) reduces loss spikes \textbf{(a)}, improves new-task accuracy \textbf{(b)}, and accelerates learning new task \textbf{(c)} compared to random initialization. A ConvNeXt pre-trained on ImageNet-1K incrementally learns 5 tasks (73 classes each) from Places-365-Std, with task switches every 1250 iterations. The backbone is frozen, and the last-layer classifier is trained using rehearsal and cross-entropy loss. In \textbf{(c)}, data-driven method reaches 95\% of random's peak accuracy with 7$\times$ fewer iterations, evaluated over 3000 iterations on the first CL task.}
   \label{fig:vis_abs}
\end{figure}

%% file: sections/2-background.tex
\section{Related Work}
\label{sec:background}


\subsection{Continual Learning}

Continual learning seeks to enable models to learn from non-stationary data while mitigating catastrophic forgetting (see \citet{zhou2023deep} for a comprehensive review). Existing CL methods can be broadly categorized into three main approaches:  
1) \textbf{Rehearsal-based (or replay-based) methods} store or reconstruct a subset of past data to replay alongside new data, thereby reducing forgetting~\citep{chaudhryER_2019, hou2019learning, rebuffi2017icarl, wu2019large}.  
2) \textbf{Regularization-based methods} introduce additional constraints in the loss function to regulate weight updates and preserve previously learned knowledge~\citep{aljundi2018memory, chaudhry2018riemannian, dhar2019learning, kirkpatrick2017overcoming}.  
3) \textbf{Parameter-isolation-based methods} allocate distinct sets of parameters or multiple model copies to different incremental tasks, preventing interference between learned representations~\citep{douillard2021dytox, yan2021dynamically, yoon2020scalable}.  
Despite these advances, most CL methods initialize the last-layer classifier with random weights, which may not be optimal when integrating new semantic categories. In this work, we primarily focus on rehearsal-based CL methods due to their superior performance~\citep{van2022three, zhou2023deep}. However, our data-driven initialization is also applicable for other CL methods.


\subsection{Weight Initialization}

Effective weight initialization stabilizes training and improves convergence. Common methods include Xavier~\citep{glorot2010understanding} and Kaiming~\citep{he2015delving}. Several approaches preserve identity mappings to enhance stability: \citet{hardt2017identity} and \citet{le2015simple} use identity mappings in residual and recurrent networks, respectively. Fixup~\citep{zhang2018fixup} and ZerO~\citep{zhao2022zero} zero out residual stems to preserve identity.
SkipInit~\citep{de2020batch} and ReZero~\citep{bachlechner2021rezero} introduce learnable parameters initialized at zero, while IDInit~\citep{pan2025idinit} uses padded identity matrices in residual networks.
These methods focus on full-network initialization in IID settings. In contrast, we study classifier weight initialization for new semantic categories in CL settings with pre-trained backbones. Prior work~\citep{harun2024overcoming, jha2024clapclip} shows class-mean initialization reduces initial training loss on new tasks. We extend this by systematically evaluating several data-driven initialization strategies, including class-mean initialization.


\subsection{Neural Collapse}
\label{sec:nc_related_work}

Neural Collapse (NC) arises when deep networks converge to highly structured representations, where class features form compact clusters that align with a simplex equiangular tight frame (Simplex ETF)~\citep{papyan2020prevalence, kothapalli2023neural, zhu2021geometric, han2022neural}. Initially observed in final-layer representations, later studies have shown that NC emerges to varying degrees in intermediate layers as well~\citep{rangamani2023feature, harun2024what, harun2025controlling}. 
Previous studies~\citep{han2022neural, zhou2022all} have shown that different loss functions e.g., cross-entropy (CE), focal loss, and MSE can lead to NC in DNNs.
NC can be characterized by four main properties:  
\begin{enumerate}[noitemsep, nolistsep, leftmargin=*]  
    \item \textbf{Feature Collapse} ($\mathcal{NC}1$):
    Features cluster tightly around the class mean, minimizing intra-class variance.
    \item \textbf{Simplex ETF Structure} ($\mathcal{NC}2$): When centered at the global mean, class means are distributed on a hypersphere with maximal pairwise distances, forming a Simplex ETF.  
    \item \textbf{Self-Duality} ($\mathcal{NC}3$): The last-layer classifiers align closely with their respective class means, creating a nearly self-dual structure.  
    \item \textbf{Nearest Class Mean Decision} ($\mathcal{NC}4$): Classification behaves like a nearest-centroid scheme. 
\end{enumerate}

%% file: sections/3-method.tex
\section{Problem Setup}
\label{formal_setting}  

We consider a CL scenario where learning begins with a pre-trained model capable of $C$-way classification. The objective is to incorporate new data—including additional classes—while preserving or improving performance on the original $C$ classes. In our study, we use an ImageNet-1K pre-trained model ($C=1000$), a common practice in CL research. However, prior work does not explicitly evaluate or preserve performance on ImageNet-1K itself~\citep{wang2022learning, wang2022dualprompt, smith2023coda, gao2023lae, mcdonnell2024ranpac}. Our work uniquely focuses on this knowledge retention aspect.  

After deployment, the pre-trained model encounters a sequence of $N-1$ tasks, $\{\mathcal{T}_2, \dots, \mathcal{T}_N\}$, where $\mathcal{T}_1$ corresponds to the pre-training data (e.g., ImageNet-1K). Each task $\mathcal{T}_j$ consists of $e_j$ labeled examples:  
$\mathcal{T}_j = \{(\mathbf{x}_i, y_i) \}_{i=1}^{e_j},$
where $\mathbf{x}_i$ is an instance of category $y_i \in Y_j$, and $Y_j$ represents the label space of task $j$. In the class-incremental learning (CIL) setting, the label spaces are disjoint, i.e., $Y_j \cap Y_{j'} = \emptyset$ for $j \neq j'$. The total number of samples across all tasks is $E = \sum_{j=1}^{N} e_j$ (ideally never ending $E \rightarrow \infty$).
During training on task $\mathcal{T}_j$, an agent has access to its current task data and any stored samples from previous tasks $\mathcal{T}_{1:j-1}$. At test time, performance is evaluated on data from all encountered classes, $\mathcal{Y}_j = Y_1 \cup \cdots \cup Y_j$, without access to task identifiers or labels.  
To ensure scalability in real-world applications, a CL system should maintain a fixed computational cost over time. For each CL task $\mathcal{T}_j$, the compute budget $\mathcal{U}$ is constrained by the number of training iterations or stochastic gradient descent steps. 
Additionally, a storage constraint $\mathcal{S}$ limits the number of samples retained in a memory buffer for rehearsal.


\section{Weight Initialization based on Least Square Weights}

We adapt the MSE loss decomposition proposed in~\citet{han2022neural} that theoretically and empirically shows how neural collapse in MSE-trained DNNs transforms the last-layer classifier into \emph{exactly} least-square classifier and the deviation from LS classifier is insignificant.
We first show the derivation of least-square weights \( \mathbf{W}_{LS} \) and describe its role in weight initialization for continual learning.  

\textbf{Deriving the Least-Square Weights.}
To understand how least-square weights arise, we begin by considering a linear classifier trained using the MSE loss. The general MSE loss for classification is defined as:
\begin{equation}
\label{eq:mse_naive}
L(\mathbf{W}, \mathbf{Z}) = \frac{1}{2} \text{Ave}_{i,c} \| \mathbf{W} \mathbf{z}_{i,c} - \mathbf{y}_{i,c} \|^2_2 + \frac{\lambda}{2} \| \mathbf{W} \|^2_F,
\end{equation}
where \( \mathbf{W} \in \mathbb{R}^{C \times (d+1)} \) is the classifier weight matrix (including bias). And, 
for the $i$-th sample in the $c$-th class, \( \mathbf{z}_{i,c} \in \mathbb{R}^{d+1} \) is the extended feature vector (including bias).  \( \mathbf{y}_{i,c} \) is the one-hot encoded target label and \( \lambda \) is the weight decay.  
The optimization objective seeks to minimize the squared error between the classifier’s predictions and the ground truth labels while imposing a weight decay penalty.
Ave is the operator that averages over its subscript indices.

\textbf{Reformulating the Loss Function.}
Rewriting the loss in matrix form, we define:  \( \mathbf{Z} \in \mathbb{R}^{(d+1) \times N} \), where columns represent feature vectors of N training samples. And, \( \mathbf{Y} \in \mathbb{R}^{C \times N} \), where columns are the corresponding one-hot labels.  
Then, the loss function in Equation~\ref{eq:mse_naive} simplifies to:
\[
L(\mathbf{W}) = \frac{1}{2N} \| \mathbf{W} \mathbf{Z} - \mathbf{Y} \|^2_2 + \frac{\lambda}{2} \| \mathbf{W} \|^2_F.
\]

\textbf{Finding the Optimal \( \mathbf{W} \) by Differentiation.}
Taking the gradient of \( L(\mathbf{W}) \) with respect to \( \mathbf{W} \) and setting it to zero:
\[
\frac{\partial L}{\partial \mathbf{W}} = \frac{1}{N} (\mathbf{W} \mathbf{Z} \mathbf{Z}^\top - \mathbf{Y} \mathbf{Z}^\top) + \lambda \mathbf{W} = 0.
\]
Solving for \( \mathbf{W} \):
\[
\mathbf{W} \mathbf{Z} \mathbf{Z}^\top + \lambda \mathbf{W} = \mathbf{Y} \mathbf{Z}^\top.
\]
Rearranging,
\[
\mathbf{W} (\mathbf{Z} \mathbf{Z}^\top + \lambda \mathbf{I}) = \mathbf{Y} \mathbf{Z}^\top.
\]
Multiplying by the inverse of \( (\mathbf{Z} \mathbf{Z}^\top + \lambda \mathbf{I}) \),
\begin{equation}
\label{eq:w_sol}
\mathbf{W} = \mathbf{Y} \mathbf{Z}^\top (\mathbf{Z} \mathbf{Z}^\top + \lambda \mathbf{I})^{-1},
\end{equation}
where \( \mathbf{I} \) is the identity matrix. 
Given, feature global mean \( \boldsymbol{\mu}_G = \text{Ave}_{i,c} \, \mathbf{z}_{i,c} \), 
feature class means \( \boldsymbol{\mu}_c = \text{Ave}_{i} \, \mathbf{z}_{i,c} \) for \( c = 1, \ldots, C \),
and feature within-class covariance \( \boldsymbol{\Sigma}_W = \text{Ave}_{i,c} \, (\mathbf{z}_{i,c} - \boldsymbol{\mu}_c)(\mathbf{z}_{i,c} - \boldsymbol{\mu}_c)^\top \), 
the total covariance matrix \( \boldsymbol{\Sigma}_T \) and class-means matrix \( \mathbf{M} \) are defined as follows:
\[
\mathbf{M} = [\boldsymbol{\mu}_1, \ldots, \boldsymbol{\mu}_C] \in \mathbb{R}^{(d+1) \times C}, \quad \boldsymbol{\Sigma}_T = \text{Ave}_{i,c} (\mathbf{z}_{i,c} - \boldsymbol{\mu}_G)(\mathbf{z}_{i,c} - \boldsymbol{\mu}_G)^\top \in \mathbb{R}^{(d+1) \times (d+1)}.
\]
We can express the solution from Equation~\ref{eq:w_sol} in terms of class statistics:
\begin{equation}
\label{eq:w_ls}
\mathbf{W}_{LS} = \frac{1}{C} \mathbf{M}^\top (\boldsymbol{\Sigma}_T + \boldsymbol{\mu}_G \boldsymbol{\mu}_G^\top + \lambda \mathbf{I})^{-1}.
\end{equation}
This formulation shows that LS weights \( \mathbf{W}_{LS} \) depend \emph{solely} on the feature statistics \( \mathbf{Z} \), making them an ideal choice for weight initialization when learning new concepts (new semantic category) in CL.
Intuitively, this method aims
to align the initial weights more closely with the distribution of the new class’s data, enabling faster and
more effective learning adjustments during the initial stage of training.

\textbf{Connecting \( \mathbf{W}_{LS} \) to Weight Initialization in Continual Learning.}  
In CIL, when a model encounters new classes, their corresponding output units are typically \emph{randomly initialized}. This causes a high initial classification loss because the newly introduced weights are unaligned with the learned feature space. As a result, the model undergoes \emph{large parameter shifts} during early training, increasing instability and leading to ``\emph{loss spikes}'' (see Fig.~\ref{fig:vis_abs}).
Instead of random initialization, we propose initializing the weights for new classes using \( \mathbf{W}_{LS} \), computed from the feature statistics of the incoming data. Since \( \mathbf{W}_{LS} \) already aligns with the feature means and covariance, it provides a \emph{low-loss starting point}, reducing the abrupt changes that occur during adaptation.
Previous studies~\citep{han2022neural, zhou2022all} have shown that DNNs trained sufficiently on a dataset converge to NC for different losses e.g., CE, MSE, etc.
Given that NC emerges regardless of the loss function used (CE or MSE), initializing new weights using the LS formulation ensures that new class units begin in a well-aligned and stable configuration.

\textbf{Practical Consideration.}
Given that CE is the most-widely used loss during pre-training stage, adapting the pre-trained models to new data with other losses e.g., MSE during CL may seem unusual. However, based on theoretical and empirical analyses, previous work found that all losses including CE, MSE achieve largely \emph{identical} features on training data by sufficiently trained DNNs~\citep{kornblith2021better, zhou2022all}. This suggests that switching loss function after pertaining stage is a reasonable consideration.

Moreover, when the pre-trained DNNs achieve neural collapse, the last-layer classifiers align
tightly with their corresponding class means, creating a
nearly self-dual configuration ($\mathcal{NC}3$)~\citep{papyan2020prevalence}. Therefore, regardless of loss functions, NC leads to desirable configuration of features and weights for data-driven weight initialization.
Additionally, we can quantify the divergence of the least-square analytical weights from the learned weights. The deviation of the analytical weights $\mathbf{W}_{LS}$ from the learned weights $\mathbf{W}$ is given by:
\begin{equation}
\label{eq:ls_deviation}
\mathcal{D}_{LS} = \frac{1}{2} tr \{ (\mathbf{W} - \mathbf{W}_{LS}) (\boldsymbol{\Sigma}_T + \boldsymbol{\mu}_G \boldsymbol{\mu}_G^\top + \lambda \mathbf{I}) (\mathbf{W} - \mathbf{W}_{LS})^\top \}.
\end{equation}
In our experiments, we find that CE-trained model achieves optimum $\mathbf{W}_{LS}$ which retains pre-training accuracy and reduces the LS deviation term $\mathcal{D}_{LS}$ (see Table~\ref{tab:loss_comp}).


\section{Loss Function}

In this work, we investigate three loss functions: CE, MSE, and Squentropy. Our objective is to find a better loss function that accelerates CL by leveraging data-driven weight initialization.
Logits $\mathbf{u}$ are defined as the matrix-vector multiplication, i.e., $\mathbf{u}= \mathbf{W}^T \mathbf{x} \in \mathbb{R}^C$ (ignoring biases), or $\mathbf{u}_i = \mathbf{w}_i \cdot \mathbf{x}$, where $\mathbf{w}_i$ is the $i$-th column of $\mathbf{W} \in \mathbb{R}^{d\times C}$ ($d$ is the embedding dimension). And, $\mathbf{t} \in \{ 0,1 \}^C$ denotes the one-hot target.

\textbf{Mean Squared Error.}
The MSE loss formulation described in Equation~\ref{eq:mse_naive} is a general form without any scaling mechanisms. Empirically, prior work has found that scaling mechanism is crucial for MSE to rival CE and improve convergence~\citep{hui2021evaluation, kornblith2021better}. Here we define a practical version of MSE with scaling parameters, $\kappa$ and $\beta$.
The MSE loss for $C-$class classification on a single input is:
\[
L_{MSE} (\mathbf{u}, \mathbf{t}, \kappa, \beta) = \frac{1}{C} \sum_{c=1}^C \Bigg( \kappa \mathbf{t}_c (\mathbf{u}_c - \beta)^2 + (1 - \mathbf{t}_c) \mathbf{u}_c^2 \Bigg ),
\]
where $\kappa$ and $\beta$ are hyperparameters. $\kappa$ weights the loss for the true class relative to incorrect classes, whereas $\beta$ controls the magnitude of the true class target. Setting $\kappa=15$ and $\beta=30$ works effectively~\citep{hui2021evaluation}.

\textbf{Cross Entropy Loss.} 
The CE loss has a term that maximizes the dot product between the logits and targets, as well as a
contractive term that minimizes the LogSumExp of the logits.
For a single input, CE loss can be defined as:
\[
L_{CE} (\mathbf{u}, \mathbf{t}) = - \frac{1}{C} \sum_{c=1}^C \mathbf{t}_c \text{ log} \left( \frac{\text{exp}(\mathbf{u}_c)}{\sum_{j=1}^C \text{exp}(\mathbf{u}_j)} \right).
\]

\textbf{Squentropy Loss.}
The squentropy loss is the sum of
two terms: the CE loss and the average
squared loss over the incorrect classes~\citep{hui2023cut}. 
Given $y$ denoting true class label, squentropy loss for a single input is defined as:
\[
L_{SQEN} (\mathbf{u}, y) = 
- \text{ log} \left( \frac{\text{exp}(\mathbf{u}_y)}{\sum_{j=1}^C \text{exp}(\mathbf{u}_j)} \right) + \frac{1}{C-1} \sum_{j=1, j \neq y}^C \mathbf{u}_j^2.
\]

%% file: sections/4-experiments.tex

\section{Experimental Setup}
\label{exp_setup}

\subsection{Dataset \& Continual Learning Setup}

We aim to study CL in an industry-like setting, which requires a large-scale data stream with numerous object categories. 
However, finding a well-curated dataset suitable for large-scale CL is challenging. 
Following~\citet{harun2024overcoming}, we construct a large-scale data stream by combining \textbf{ImageNet-1K} (1.2 million images) with \textbf{Places-365-Standard} (1.8 million images). Places-365 is a challenging dataset~\citep{liu2019large} widely used for out-of-distribution detection with ImageNet-pretrained models~\citep{zhu2022boosting}.  
This combined dataset, consisting of 3 million images and 1365 classes, allows us to study the impact of weight initialization in large-scale CL.

During CIL, the model sequentially learns $5$ disjoint tasks from Places-365, each of which contains 73 non-overlapping classes.  
Rehearsal updates the model over 600 iterations per task, where each iteration processes 256 samples—50\% randomly selected from the current task and 50\% from previously seen tasks and ImageNet-1K.  
To isolate the effect of weight initialization, augmentation is omitted. 
Performance is evaluated every 50 training iterations using a test set comprising the ImageNet-1K classes and all encountered Places-365 classes from both current and previous CL tasks. We set $\lambda$ to 0.05 for computing LS weights $\mathbf{W}_{LS}$ (Eq.~\ref{eq:w_ls}), based on the weight decay parameter.

\textbf{Loss Alignment.}  
When using MSE loss which is different from the pre-training loss i.e., CE, we perform a \textit{loss alignment fine-tuning phase} before CL begins.  
We train the last-layer classifier on ImageNet-1K dataset using MSE loss, while keeping the backbone frozen.
We include dataset and implementation details in Appendix~\ref{sce:details} and~\ref{sec:implement}.

\subsection{Compute \& Storage Constraints}  
A continual learner must adapt to a large-scale data stream without incurring an increasing computational burden over time. 
Recent works have emphasized computational efficiency in CL~\citep{prabhu2023computationally, harun2023siesta, harun2023efficient, harun2024grasp, verwimp2023continual, zhang2023continual}.  
In our experiments, we constrain computation by fixing the number of training iterations or gradient descent steps. In particular, we bound compute by 1200 or 600 training iterations.  
We also impose a fixed replay buffer size to limit storage usage. During learning, the model rehearses samples from the buffer, which is capped at a maximum of 192K or 24K samples—equivalent to 6.4\% and 0.8\% of the entire dataset (3 million samples from ImageNet and Places combined), respectively. For buffer maintenance, we adopt a class-balanced policy by storing an equal number of randomly selected samples per class.


\subsection{Architecture}  
We select the ConvNeXt architecture due to its modern design and superior performance compared to similar-sized DNNs.  
Our main experiments use \textbf{ConvNeXtV2-Femto}~\citep{woo2023convnext}, pretrained on ImageNet-1K using a fully convolutional masked autoencoder framework, followed by supervised fine-tuning.  
Although ResNet18~\citep{he2016deep} is commonly used in CL, it underperforms relative to other lightweight DNNs~\citep{hayes2022online, harun2023siesta}.  
ConvNeXtV2-Femto has $5.2$M parameters—2$\times$ fewer than ResNet18's $11.6$M—while achieving an absolute 8.47\% higher top-1 accuracy on ImageNet-1K. 
Furthermore, unlike ResNet, ConvNeXt is amenable to parameter-efficient fine-tuning (PEFT) approaches e.g., Low-Rank-Adaptation (LoRA)~\citep{hu2022lora} due to the linear layers. Additionally, we conduct ablation experiments with \textbf{ResNet18} in Sec.~\ref{sec:ablation}.

\subsection{Controlling Plasticity}
\label{sec:control_plasticity_setting}

The rise of large foundation models has spurred interest in integrating CL with pre-trained models. 
However, naively fine-tuning pre-trained models significantly degrades CL performance. 
Prior work shows that \emph{controlling plasticity}—selectively updating parts of the model—is key to effective adaptation~\citep{harun2024overcoming}.
In this work, we explore two approaches for controlling plasticity in pre-trained backbones:
\begin{enumerate}[noitemsep, nolistsep, leftmargin=*]
    \item \textbf{CL with frozen backbone}: 
    Here we ask: \emph{how can we adapt a pre-trained model without updating its representations?}
    We train the last-layer classifier while keeping the backbone frozen, treating it as a fixed feature extractor.
    
    \item \textbf{CL with controlled plasticity}: 
    Here we ask: \emph{how can we selectively update the representations of a pre-trained model to incorporate new knowledge?}
    Inspired by prior work~\citep{harun2024overcoming}, we fine-tune the top layers of the pre-trained backbone using LoRA. During CL, only the LoRA parameters are updated with rehearsal, while the original backbone weights remain frozen. After rehearsal, the LoRA weights are merged with the backbone, and the last-layer classifier is updated as usual. 
    In particular, we keep top two ConvNeXt blocks plastic (i.e., 55\% of the parameters) and keep remaining blocks frozen. Within each block, two linear layers are modified to incorporate LoRA’s weights with rank 48. To initialize LoRA weights, we adhere to LoRA paper~\citep{hu2022lora}. 
    Additional implementation details are provided in Appendix~\ref{sec:lora}.
    
\end{enumerate}

\subsection{Evaluation Criteria}

Motivated by prior work~\citep{harun2024overcoming, koh2022online}, we perform continual evaluation or any-time-inference. 
We compute \textbf{average accuracy}~\citep{koh2022online}, defined as:  
$\mathcal{A}^{avg} = \frac{1}{N} \sum_{i=1}^{N} \mathcal{A}_{i},$
where $N$ is the total number of evaluation steps, and $\mathcal{A}_{i}$ represents top-1 accuracy (\%) at step $i$.  
To analyze CL performance, we report:  
\begin{itemize}[noitemsep, nolistsep, leftmargin=*]  
    \item \textbf{Plasticity:} $\mathcal{A}_{new}$—performance on new (or current) task.
    \item \textbf{Stability:} $\mathcal{A}_{old}$—performance on previously learned tasks.  
    \item \textbf{Overall CL performance:} $\mathcal{A}_{all}$—accuracy across all encountered tasks including new and old tasks.
    \item \textbf{Pre-training task performance:} $\mathcal{A}_{pre}$—performance on the initial ImageNet-1K pre-training task.
    \item \textbf{Forward transfer:} $\mathcal{A}_\text{fw}$—accuracy evaluated on future task using linear probing. 
    Details are in Appendix~\ref{sec:linear_probing}.
    \item \textbf{Training loss on new task:}
    $\mathcal{L}_{new}$—average training loss over evaluation steps for new task. 
    \item \textbf{Efficiency gain:} $\mathcal{G}$—average efficiency gain or speedup in learning new tasks. Details are given in Appendix~\ref{sec:eff_gain}.
\end{itemize}






\section{Experimental Results}
\label{sec:results}


We evaluate three weight initialization methods: (a) \textbf{random initialization}, (b) \textbf{class-mean initialization}, where the weight $\mathbf{w}_c$ for class $c$ is set to the class mean $\boldsymbol{\mu}_c$, computed as $\boldsymbol{\mu}_c = \text{Ave}_{i} \, \mathbf{z}_{i,c}$, and (c) \textbf{least-square initialization}, where the weight $\mathbf{W}_{LS}$ is derived using both buffered old data and new data, as defined in Equation~\ref{eq:w_ls}.


We present the results as follows: Sec.~\ref{sec:init_first_task} analyzes the impact of weight initialization on the first CL task. Sec.~\ref{sec:frozen_backbone} and Sec.~\ref{sec:control_plasticity} evaluate weight initialization under frozen backbone and controlled plasticity settings. Sec.~\ref{sec:init_cl_methods} examines the effectiveness of data-driven initialization with various CL methods, and Sec.~\ref{sec:ablation} provides additional ablation results.

\subsection{Impact of Weight Initialization on the First CL Task}
\label{sec:init_first_task}



We first assess the impact of weight initialization on the first CL task before any training occurs. Using an ImageNet-1K pre-trained ConvNeXt, we augment the last-layer classifier with new class units initialized using random, LS, and class-mean methods. The results are summarized in Table~\ref{tab:loss_comp}.
Surprisingly, CE loss yields a lower LS deviation ($4.78\times$ lower $\mathcal{D}_{LS}$) than MSE loss. While LS initialization should theoretically favor MSE due to its connection with MSE loss decomposition, our results suggest that CE-trained features better align with LS weights. This indicates that NC strongly shapes the organization of penultimate-layer features and last-layer weights, regardless of the loss function.

\input{tables/weight_drift}

\input{tables/last_layer_plastic}

Furthermore, NC evaluations ($\mathcal{NC}1=0.32, \mathcal{NC}2=0.93, \mathcal{NC}3=0.90, \mathcal{NC}4=1.59$) confirm that CE-learned features exhibit stronger neural collapse. A lower $\mathcal{NC}$ value indicates stronger collapse. These results support our arguments on using LS weight initialization for CE or other losses as long as pre-training ensures NC criteria.
As shown in Table~\ref{tab:loss_comp}, random initialization fails to achieve \emph{nonzero} accuracy on new tasks, since it does not encode any feature statistics. Conversely, least-square yields the highest accuracy, followed by class-mean.
Efficiency gain is illustrated in Fig.~\ref{fig:eff_gain}.
These results suggest that data-driven initialization provides a strong starting point, enabling continual learners to adapt quickly to new tasks.



\subsection{CL with Frozen Backbone}
\label{sec:frozen_backbone}

We next analyze the impact of weight initialization on the last-layer classifier under a frozen backbone setting. The results are summarized in Table~\ref{tab:output_layer}.
Compared to random initialization, LS improves new task accuracy (plasticity) by an absolute margin of $4.55\%-16.25\%$, while maintaining stability. Class-mean initialization also provides improvements, with gains of $2.29\%-16.35\%$. Overall, LS outperforms class-mean initialization. Interestingly, LS benefits CE and squentropy losses more than MSE, suggesting that LS initialization requires an MSE-trained backbone to reach its full potential. Due to computational constraints, our loss alignment experiments fine-tune only the last-layer classifier on ImageNet-1K; full-network fine-tuning would likely enhance LS benefits and alignment.
In terms of training loss, LS consistently achieves the lowest loss across all experiments. 
Appendix~\ref{sec:more_results} includes qualitative analysis.
Among loss functions, CE delivers the highest performance, followed by squentropy.

\textbf{Learning Efficiency.} 
We also evaluate the efficiency of learning new tasks. LS initialization provides a speedup of $3.74\times - 5.29\times$, while class-mean initialization offers $2\times - 3.74\times$ speedup. Our results suggest that task-aware initialization mitigates loss spikes and improves learning efficiency. Among loss functions, CE converges the fastest.

\input{tables/top_layers_plastic}


\subsection{CL with Controlled Plasticity}
\label{sec:control_plasticity}


To assess the impact of weight initialization on representation learning, we train only the top layers of the network. Prior work~\citep{harun2024overcoming} emphasizes the importance of controlling plasticity to integrate new concepts without forgetting prior knowledge. Following their approach, we use LoRA to regulate plasticity in the top layers. Results in Table~\ref{tab:top_layers} show that updating these layers (vs. only the last layer; see Sec.~\ref{sec:frozen_backbone}) improves CL performance, suggesting that selective updates help integrate new concepts.

LS initialization achieves the highest new-task accuracy in most settings. Consistent with the frozen-backbone setting, LS struggles with MSE loss due to loss alignment issues but performs well with CE and squentropy losses, improving new-task accuracy by an absolute margin of $6.58\%-8.98\%$ compared to random baseline. Additionally, LS consistently reduces new-task loss across all comparisons and achieves higher accuracy on all tasks than other initializations.
Class-mean initialization also proves effective but generally underperforms LS.
Furthermore, LS and class-mean initializations improve \emph{forward transfer}—where learning the current task benefits future tasks—making them important for accelerating CL. This highlights the role of data-driven initialization in facilitating transferable representations during CL. Among the studied loss functions, CE consistently achieves the best CL performance across all metrics.

\textbf{Learning Efficiency.} 
LS and class-mean initializations speed up training by up to $4.77\times$ over random, highlighting their efficiency and effectiveness in improving CL performance across loss functions.

\subsection{Impact of Weight Initialization on CL Methods}
\label{sec:init_cl_methods}

We investigate the following question: \emph{does the least-square initialization remain effective when integrated with different CL methods?} To evaluate this, we consider two widely studied CL methods: \textbf{EWC}~\citep{kirkpatrick2017overcoming} for parameter regularization and \textbf{DER++}~\citep{buzzega2020dark} for rehearsal combined with distillation. This experiment follows the CL with controlled plasticity setting described in Sec.\ref{sec:control_plasticity_setting}. 
A ConvNeXtV2-F model, pre-trained on ImageNet-1K, incrementally learns five CIL tasks, each comprising 73 classes from Places365-Std. Further implementation details are given in Appendix~\ref{sec:ablation_exp}. Results are summarized in Table~\ref{tab:cl_methods}, where we find that the least-square initialization consistently outperforms both random and class-mean initializations across all CL methods, highlighting its robustness and general effectiveness. 
Notably, least-square initialization outperforms random initialization in new task accuracy by an absolute margin of 13.19\% for EWC and 8.45\% for DER++, demonstrating its efficacy in mitigating loss spikes and enabling more efficient adaptation.

\input{tables/cl_methods_exp} 

\input{tables/last_layer_plastic_resnet}

\subsection{Additional Experiments}
\label{sec:ablation}
We summarize additional supporting results and provide implementation details in Appendix~\ref{sec:implement}.

\textbf{ResNet Experiments.}
Besides ConvNeXt architecture, we also experiment with the widely used ResNet18 architecture. Following the frozen backbone setting described in Sec.~\ref{sec:control_plasticity_setting}, we incrementally train ImageNet-1K pre-trained ResNet18 models on five CIL tasks from Places365-Std (73 classes per CL task) using CE loss. As shown in Table~\ref{tab:output_layer_resnet}, least square consistently outperforms random and class-mean baselines in all criteria, demonstrating efficacy for ResNet18 architecture. Notably, least-square exceeds random initialization by a 12.86--13.62\% absolute margin in new task accuracy, $\mathcal{A}_{\text{new}}$.
It reduces training loss ($\mathcal{L}_{\text{new}}$) by a factor of 3.6--5.4$\times$ compared to random initialization, showing significant impact of data-driven weight initialization on mitigating loss spikes.

\textbf{CIL with Multiple Data Orderings.}
To evaluate the robustness of our findings, we extend our experiments to multiple data orderings.  
For each weight initialization method, we conduct five CIL experiments using five random seeds and class orders. In each experiment, an ImageNet-1K pretrained ConvNeXtV2-F model incrementally learns five disjoint tasks from Places365-Std (73 classes per task) using CE loss. These experiments follow the CL with controlled plasticity setting described in Sec.\ref{sec:control_plasticity_setting}. 
The computational and storage budgets are constrained to 600 iterations and 24K samples.
The averaged results across five data orderings are given in Table~\ref{tab:dataorders}. We find that least-square consistently yields lower loss and higher accuracy than competing methods, demonstrating robust performance.

\input{tables/exp_multiseed}

\textbf{Computational Overhead.}
Least-square initialization requires only a single additional forward pass to compute the global mean, class mean matrix, and covariance matrix (Equation~\ref{eq:w_ls}), using a small subset of buffered old data and new data. 
As shown in Table~\ref{tab:time} (Appendix), the added cost of least-square initialization in terms of wall-clock runtime is 0.07\%—a trivial overhead that we believe is well justified given the observed performance gains.


%% file: tables/weight_drift.tex
\begin{table}[t]
\centering
  \caption{\textbf{Weight initializations in the first CL task.} We adapt an ImageNet-1K pre-trained model to the first CL task, i.e., 73 classes from Places-365-Standard dataset.
  $\mathcal{D}_{LS}$ denotes the deviation of the least-square analytical weights $\mathbf{W}_{LS}$ from the learned weights $\mathbf{W}$, which correspond to ImageNet-1K pre-training (see Equation~\ref{eq:ls_deviation} for the definition). Accuracy is reported in \%. Arrows indicate the optimal direction of performance. 
  }
  \label{tab:loss_comp}
  \centering
     \begin{tabular}{cccccc}
     \toprule
     \multicolumn{1}{c}{\textbf{Weight Init}} &
     \multicolumn{1}{c}{$\boldsymbol{\mathcal{A}}_{\text{pre}} \uparrow$} &
     \multicolumn{1}{c}{$\boldsymbol{\mathcal{A}}_{\text{new}} \uparrow$} &
     \multicolumn{1}{c}{$\boldsymbol{\mathcal{A}}_{\text{old}} \uparrow$} &
     \multicolumn{1}{c}{$\boldsymbol{\mathcal{A}}_{\text{all}} \uparrow$} &
     \multicolumn{1}{c}{$\boldsymbol{\mathcal{D}}_{LS} \downarrow$} \\
     \midrule
     \textcolor{orange}{\textbf{MSE Loss}} \\
     Random &  & 0.00 & 77.72 & 67.82 &  \\
     Class Mean & 77.72 & 50.29 & 68.34 & 66.04 & 377.18 \\
     \textbf{Least Square} &  & \textbf{51.70} & 71.57 & \textbf{69.04} &  \\
     \midrule
     \textcolor{orange}{\textbf{CE Loss}} \\
     Random &  & 0.00 & 78.23 & 68.26 &  \\
     Class Mean & 78.23 & 21.10 & 77.87 & 70.64 & \textbf{78.97} \\
     \textbf{Least Square} &  & \textbf{48.37} & 75.83 & \textbf{72.33} &  \\
    \bottomrule
    \end{tabular}
\end{table}

%% file: tables/last_layer_plastic.tex
\begin{table}[ht]
\centering
  \caption{\textbf{Last-Layer Plastic.} 
  An ImageNet-1K pre-trained model incrementally learns 5 disjoint tasks, each consisting of 73 classes from Places-365-Std. 
  $\mathcal{U}$ and $\mathcal{S}$ denote compute and storage constraints respectively. 
  $\mathcal{G}$ denotes average efficiency gain in learning new tasks. 
  Forward transfer ($\mathcal{A}_\text{fw}$) is not applicable for this setting since the backbone is frozen and identical for all methods. Accuracy is reported in \%. Arrows indicate the optimal direction of performance.
  }
  \label{tab:output_layer}
  \centering
     \begin{tabular}{cccccccccc}
     \toprule
     \multicolumn{1}{c}{\textbf{Weight Init}} &
     \multicolumn{1}{c}{\textbf{Loss}} &
     \multicolumn{1}{c}{$\boldsymbol{\mathcal{U}}$} &
     \multicolumn{1}{c}{$\boldsymbol{\mathcal{S}}$} &
     \multicolumn{1}{c}{$\boldsymbol{\mathcal{L}}_{\text{new}} \downarrow$} &
     \multicolumn{1}{c}{$\boldsymbol{\mathcal{A}}_{\text{pre}} \uparrow$} &
     \multicolumn{1}{c}{$\boldsymbol{\mathcal{A}}_{\text{new}} \uparrow$} &
     \multicolumn{1}{c}{$\boldsymbol{\mathcal{A}}_{\text{old}} \uparrow$} &
     \multicolumn{1}{c}{$\boldsymbol{\mathcal{A}}_{\text{all}} \uparrow$} &
     \multicolumn{1}{c}{$\boldsymbol{\mathcal{G}} \uparrow$} \\
    \midrule
    Random &  & & & 2.30 & 75.52 & 49.70 & 65.85 & 64.10 & 1.00 \\
    Class Mean & CE & 1200 & 192K & 1.65 & 75.61 & 54.64 & 65.86 & 64.60 & 3.08 \\
    \textbf{Least Square} &  &  &  & \textbf{1.50} & 75.52 & \textbf{57.28} & 65.91 & \textbf{64.91} & \textbf{5.29} \\
    \hline
    Random &  & & & 2.47 & 74.85 & 52.73 & 63.95 & 62.67 & 1.00 \\
    Class Mean & SQEN & 1200 & 192K & 2.35 & 74.77 & 55.02 & 63.81 & 62.78 & 2.72 \\
    \textbf{Least Square} &  &  &  & \textbf{2.24} & 74.63 & \textbf{57.28} & 63.77 & \textbf{62.98} & \textbf{4.84} \\
    \hline
    Random &  & & & 4.78 & 72.80 & 44.63 & 60.77 & \textbf{58.96} & 1.00 \\
    Class Mean & MSE & 1200 & 192K & 6.23 & 70.11 & \textbf{53.68} & 57.25 & 56.73 & 2.22 \\
    \textbf{Least Square} &  &  &  & \textbf{4.11} & 71.45 & 53.26 & 59.41 & 58.67 & \textbf{4.17} \\
    \toprule
    Random &  &  &  & 2.79 & 74.12 & 43.28 & 64.48 & 62.21 & 1.00 \\
    Class Mean & CE & 600 & 24K & 1.71 & 74.18 & 51.82 & 64.32 & 62.91 & 2.68 \\
    \textbf{Least Square} &  &  &  & \textbf{1.58} & 73.95 & \textbf{55.29} & 64.34 & \textbf{63.29} & \textbf{3.80} \\
    \hline
    Random &  &  &  & 2.71 & 74.09 & 47.63 & 63.27 & 61.52 & 1.00 \\
    Class Mean & SQEN & 600 & 24K & 2.47 & 73.93 & 52.08 & 62.99 & 61.73 & 2.01 \\
    \textbf{Least Square} &  &  &  & \textbf{2.34} & 73.62 & \textbf{55.58} & 62.91 & \textbf{62.03} & \textbf{3.80} \\
    \hline
    Random &  &  &  & 5.57 & 73.46 & 35.43 & 62.00 & \textbf{59.10} & 1.00 \\
    Class Mean & MSE & 600 & 24K & 8.34 & 68.52 & \textbf{51.78} & 55.83 & 55.26 & \textbf{3.74} \\
    \textbf{Least Square} &  &  &  & \textbf{4.41} & 70.69 & 51.68 & 59.45 & 58.56 & 3.56 \\
    \bottomrule
    \end{tabular}
\end{table}

%% file: tables/top_layers_plastic.tex
\begin{table}[ht]
\centering
  \caption{\textbf{Top Layers Plastic.} 
  An ImageNet-1K pre-trained model incrementally learns 5 disjoint tasks, each consisting of 73 classes from Places-365-Std.
  $\mathcal{U}$ and $\mathcal{S}$ denote compute and storage constraints respectively. 
  $\mathcal{G}$ denotes average efficiency gain in learning new tasks. $\mathcal{A}_\text{fw}$ denotes forward transfer accuracy (\%) evaluated on future task.
  Accuracy is reported in \%. Arrows indicate the optimal direction of performance.
  }
  \label{tab:top_layers}
  \centering
     \begin{tabular}{ccccccccccc}
     \toprule
     \multicolumn{1}{c}{\textbf{Weight Init}} &
     \multicolumn{1}{c}{\textbf{Loss}} &
     \multicolumn{1}{c}{$\boldsymbol{\mathcal{U}}$} &
     \multicolumn{1}{c}{$\boldsymbol{\mathcal{S}}$} &
     \multicolumn{1}{c}{$\boldsymbol{\mathcal{L}}_{\text{new}} \downarrow$} &
     \multicolumn{1}{c}{$\boldsymbol{\mathcal{A}}_{\text{pre}} \uparrow$} &
     \multicolumn{1}{c}{$\boldsymbol{\mathcal{A}}_{\text{new}} \uparrow$} &
     \multicolumn{1}{c}{$\boldsymbol{\mathcal{A}}_{\text{old}} \uparrow$} &
     \multicolumn{1}{c}{$\boldsymbol{\mathcal{A}}_{\text{all}} \uparrow$} &
     {$\boldsymbol{\mathcal{A}}_{\text{fw}} \uparrow$} &
     \multicolumn{1}{c}{$\boldsymbol{\mathcal{G}} \uparrow$} \\
    \midrule
    Random &  & & & 2.28 & 77.47 & 51.70 & 68.43 & 66.58 & 67.69 & 1.00 \\
    Class Mean & CE & 1200 & 192K & 1.58 & 77.47 & 58.90 & 68.11 & 67.04 & 68.00 & 2.03 \\
    \textbf{Least Square} &  &  &  & \textbf{1.55} & 76.94 & \textbf{59.60} & 68.11 & \textbf{67.13} & \textbf{68.04} & \textbf{2.31} \\
    \hline
    Random &  & & & 2.55 & 76.22 & 52.39 & 65.96 & 64.42 & 65.66 & 1.00 \\
    Class Mean & SQEN & 1200 & 192K & 2.07 & 76.06 & 58.85 & 65.68 & 64.84 & 66.63 & 2.54 \\
    \textbf{Least Square} &  &  &  & \textbf{2.04} & 75.96 & \textbf{58.97} & 65.84 & \textbf{65.01} & \textbf{66.76} & \textbf{2.72} \\
    \hline
    Random &  & & & 4.38 & 74.82 & 40.39 & 61.89 & \textbf{59.52} & 58.67 & 1.00 \\
    Class Mean & MSE & 1200 & 192K & 4.03 & 70.22 & 55.25 & 57.66 & 57.27 & \textbf{64.61} & 3.53 \\
    \textbf{Least Square} &  &  &  & \textbf{3.41} & 71.61 & \textbf{54.67} & 59.66 & 59.04 & 63.45 & \textbf{3.95} \\
    \toprule
    Random &  & & & 2.40 & 74.77 & 48.72 & 65.64 & 63.76 & 65.58 & 1.00 \\
    Class Mean & CE & 600 & 24K & 1.44 & 74.46 & 56.49 & 65.43 & 64.39 & 66.74 & 2.16 \\
    \textbf{Least Square} &  &  &  & \textbf{1.39} & 74.35 & \textbf{57.70} & 65.69 & \textbf{64.77} & \textbf{66.80} & \textbf{2.98} \\
    \hline
    Random &  & & & 2.76 & 75.28 & 48.01 & 64.88 & 63.00 & 63.77 & 1.00 \\
    Class Mean & SQEN & 600 & 24K & 2.16 & 75.02 & 56.30 & 64.56 & 63.57 & 64.84 & 2.13 \\
    \textbf{Least Square} &  &  &  & \textbf{2.12} & 74.86 & \textbf{56.65} & 64.87 & \textbf{63.91} & \textbf{65.03} & \textbf{2.60} \\
    \hline
    Random &  & & & 5.0 & 74.65 & 33.26 & 61.70 & \textbf{58.60} & 56.58 & 1.00 \\
    Class Mean & MSE & 600 & 24K & 4.32 & 69.19 & \textbf{53.18} & 57.21 & 56.66 & \textbf{62.88} & \textbf{4.77} \\
    \textbf{Least Square} &  &  &  & \textbf{3.77} & 70.48 & 52.15 & 59.06 & 58.26 & 62.39 & 4.55 \\
    \bottomrule
    \end{tabular}
\end{table}

%% file: tables/cl_methods_exp.tex
\begin{table}[t]
\centering
\caption{\textbf{Impact of Weight Initialization on CL Methods.} An ImageNet-1K pre-trained ConvNeXtV2-F is incrementally trained on 5 CIL tasks from Places-365-Std (73 classes per task). Rehearsal is used for DER++ but omitted for EWC. For both methods, compute is bounded by 600 iterations. For DER++, buffer is constrained by 24K samples. Accuracy is reported in \%. Arrows indicate the optimal direction of performance. 
}
\label{tab:cl_methods}
\begin{tabular}{c|cccc|cccc}
\toprule
\textbf{Weight Init} &
\multicolumn{4}{c|}{\textbf{EWC}} &
\multicolumn{4}{c}{\textbf{DER++}} \\
& 
$\boldsymbol{\mathcal{L}}_{\text{new}} \downarrow$ & 
$\boldsymbol{\mathcal{A}}_{\text{new}} \uparrow$ &  
$\boldsymbol{\mathcal{A}}_{\text{old}} \uparrow$ & 
$\boldsymbol{\mathcal{A}}_{\text{all}} \uparrow$ &  
$\boldsymbol{\mathcal{L}}_{\text{new}} \downarrow$ & 
$\boldsymbol{\mathcal{A}}_{\text{new}} \uparrow$ & 
$\boldsymbol{\mathcal{A}}_{\text{old}} \uparrow$ & 
$\boldsymbol{\mathcal{A}}_{\text{all}} \uparrow$ \\
\midrule
Random & 2.06 & 47.46 & 41.83 & 42.18 & 2.15 & 50.25 & 64.86 & 63.21 \\
Class Mean & 1.36 & 59.38 & 39.59 & 41.38 & 1.36 & 58.22 & 64.64 & 63.84 \\
\textbf{Least Square} & \textbf{1.28} & \textbf{60.65} & 40.74 & \textbf{42.57} & \textbf{1.34} & \textbf{58.70} & 65.00 & \textbf{64.23} \\
\bottomrule
\end{tabular}
\end{table}

%% file: tables/last_layer_plastic_resnet.tex
\begin{table}[t]
\centering
  \caption{\textbf{ResNet Results.} 
  An ImageNet-1K pre-trained ResNet18 incrementally learns 5 disjoint tasks (73 Places classes per task) using CE loss and last-layer plastic setting. All metrics except $\mathcal{L}_{\text{new}}$ are reported in \%. $\mathcal{U}$ and $\mathcal{S}$ denote compute and storage constraints respectively. Arrows indicate the optimal direction of performance. 
  }
  \label{tab:output_layer_resnet}
  \centering
     \begin{tabular}{cccccccc}
     \toprule
     \multicolumn{1}{c}{\textbf{Weight Init}} &
     \multicolumn{1}{c}{$\boldsymbol{\mathcal{U}}$} &
     \multicolumn{1}{c}{$\boldsymbol{\mathcal{S}}$} &
     \multicolumn{1}{c}{$\boldsymbol{\mathcal{L}}_{\text{new}} \downarrow$} &
     \multicolumn{1}{c}{$\boldsymbol{\mathcal{A}}_{\text{pre}} \uparrow$} &
     \multicolumn{1}{c}{$\boldsymbol{\mathcal{A}}_{\text{new}} \uparrow$} &
     \multicolumn{1}{c}{$\boldsymbol{\mathcal{A}}_{\text{old}} \uparrow$} &
     \multicolumn{1}{c}{$\boldsymbol{\mathcal{A}}_{\text{all}} \uparrow$} \\
    \midrule
    Random &  & & 4.82 & 66.38 & 10.79 & 54.55 & 49.95  \\
    Class Mean & 1200 & 192K & 1.69 & 65.61 & 21.17 & 54.23 & 50.70 \\
    \textbf{Least Square} &  &  & \textbf{1.34} & 65.96 & \textbf{24.41} & \textbf{57.39} & \textbf{53.94} \\
    \midrule
    Random &  &  & 6.18 & 65.85 & 3.60 & 52.90 & 47.68  \\
    Class Mean & 600 & 24K & 1.44 & 64.32 & 12.08 & 51.97 & 47.71 \\
    \textbf{Least Square} &  &  & \textbf{1.15} & 65.05 & \textbf{16.46} & \textbf{56.34} & \textbf{52.18} \\
    \bottomrule
    \end{tabular}
\end{table}

%% file: tables/exp_multiseed.tex
\begin{table}[t]
\centering
  \caption{\textbf{Results with Multiple Data Orderings.} 
  Results after learning ImageNet-1K followed by Places-365-Standard over 5 CIL tasks, each containing 73 classes. Each weight initialization method is evaluated using five different class orderings and random seeds. CE loss is used for training. The computational and storage budgets are constrained to 600 iterations and 24K samples. We report the average accuracy (in \%) and standard deviation across five runs. Arrows indicate the optimal direction of performance. 
  }
  \label{tab:dataorders}
  \centering
     \begin{tabular}{cccccc}
     \toprule
     \textbf{Weight Init} & $\boldsymbol{\mathcal{L}}_{\text{new}} \downarrow$ & $\boldsymbol{\mathcal{A}}_{\text{pre}} \uparrow$ & $\boldsymbol{\mathcal{A}}_{\text{new}} \uparrow$ & $\boldsymbol{\mathcal{A}}_{\text{old}} \uparrow$ & $\boldsymbol{\mathcal{A}}_{\text{all}} \uparrow$  \\
    \midrule
     Random & $2.38$ \scriptsize $\pm0.02$ & $74.83$ \scriptsize $\pm0.06$ & $49.34$ \scriptsize $\pm0.40$ & $65.59$ \scriptsize $\pm0.10$ & $63.79$ \scriptsize $\pm0.09$ \\
     Class Mean & $1.43$ \scriptsize $\pm0.02$ & $74.49$ \scriptsize $\pm0.04$ & $57.23$ \scriptsize $\pm0.45$ & $65.37$ \scriptsize $\pm0.10$ & $64.42$ \scriptsize $\pm0.09$ \\
     \textbf{Least Square} & $\mathbf{1.39}$ \scriptsize $\pm0.01$ & $74.33$ \scriptsize $\pm0.02$ & $\mathbf{58.29}$ \scriptsize $\pm0.39$ & $65.56$ \scriptsize $\pm0.12$ & $\mathbf{64.72}$ \scriptsize $\pm0.10$ \\
    \bottomrule
    \end{tabular}
\end{table}

%% file: sections/5-discussion.tex
\section{Discussion}
\label{sec:discussion}
We show that least-square initialization outperforms random initialization in large-scale CL by aligning classifier weights with feature statistics, reducing initial instability and enabling smoother task adaptation. This method integrates seamlessly into existing CL frameworks without modifying pre-trained networks or adding memory overhead.
With the rising energy demands of AI~\citep{luccioni2022estimating, patterson2021carbon, wu2022sustainable}, CL has the potential to reduce carbon emissions. Despite progress, most CL methods offer little computational efficiency gains~\citep{harun2023efficient}.
While PEFT (e.g., LoRA) improves efficiency, we demonstrate that data-driven weight initialization further boosts both efficiency and performance in CL under distribution shifts.

\textbf{\emph{Future Directions.}}
While data-driven initialization improves CL, several open questions remain. Our approach is formulated under the MSE loss framework. Investigating whether similar principles hold under alternative objectives, such as contrastive or meta-learning losses, is an exciting avenue for future research.
Additionally, our study focuses on image classification tasks. Extending our approach to other tasks, such as object detection~\citep{acharya2020rodeo} and language understanding~\citep{jangtowards}, could provide further insights into its generalizability.
Due to computational limits, we evaluated up to 1365 classes; future work should test scalability to larger, more diverse datasets.

Alternative loss functions that address the limitations of CE loss (e.g., contrastive term between classes) could further accelerate CL.
While MSE loss offers promise, its performance depends heavily on heuristically tuned scaling parameters, critical for optimization and neural collapse~\citep{hui2021evaluation,
hui2023cut, zhou2022optimization}. The optimal design of these parameters remains an open theoretical challenge. 
When designing loss functions for CL, \emph{forward transfer} must be considered, as several loss functions—including MSE—that outperform CE on in-distribution data often exhibit poor transferability to out-of-distribution datasets~\citep{kornblith2021better, harun2024what, harun2025controlling}.


%% file: sections/6-conclusion.tex
\section{Conclusion}
\label{sec:conclusion}

By integrating theoretical insights from neural collapse with practical CL improvements, our work offers a new perspective on data-driven weight initialization. We demonstrate that least-square weight initialization effectively prevents loss spikes and enables faster adaptation in CL.
We believe that data-driven weight initialization can enhance learning efficiency and strengthen CL in addressing real-world challenges.

%% file: sections/supplemental.tex
\clearpage
\setcounter{page}{1}

\appendix

\begin{center}
    {\Large{Appendix}}
\end{center}


We organize Appendix as follows: 
\begin{itemize}
    \item Appendix~\ref{sce:details} provides details on the datasets used in this paper.
    \item Appendix~\ref{sec:implement} provides additional implementation and training details.
    \item Appendix~\ref{sec:nc_metrics} describes neural collapse metrics ($\mathcal{NC}1$-$\mathcal{NC}4$).
    \item Appendix~\ref{sec:more_results} includes additional experimental results.
\end{itemize}

In this paper, we use several acronyms such as
\textbf{CL} : Continual Learning,
\textbf{DNN} : Deep Neural Network,
\textbf{CIL} : Class Incremental Learning,
\textbf{IID} : Independent and Identically Distributed,
\textbf{NC} : Neural Collapse, 
\textbf{ETF} : Equiangular Tight Frame,
\textbf{LS} : Least-Square,
\textbf{CE} : Cross Entropy, 
\textbf{MSE} : Mean Squared Error,
\textbf{SQEN} : Squentropy.

We used a single NVIDIA RTX A5000 GPU to run each CL experiment.

\section{Dataset Details}
\label{sce:details}

In this work, we use two large-scale datasets namely \textbf{ImageNet-1K} and \textbf{Places-365-Standard} to construct a large-scale data stream for CL.

ImageNet-1K~\citep{russakovsky2015imagenet} is the standard object recognition benchmark for testing a model’s ability to scale. It has over $1.28$ million images uniformly distributed across $1000$ categories. Each object category consists of $732-1300$ training images and $50$ validation images.

Places-365-Standard is a subset of Places-2 Dataset~\citep{zhou2017places}.
Places-365-Standard~\citep{zhou2017places} has over $1.8$ million training images from $365$ different scene
categories with $3068-5000$ images per class. Each image depicts a specific scene or environment. The images in the dataset have $256 \times 256$ pixels.
We use the validation set consisting of $100$ images per class to test the models.

In all experiment, the input image resolution is $224 \times 224$. In both training and test time, images are pre-processed by first resizing to $256 \times 256$ and center cropping with a size of $224 \times 224$.

\section{Implementation Details}
\label{sec:implement}

In this section, we provide additional implementation details. In all experiments, we use ConvNeXtV2 backbone~\footnote{Pre-trained weights are available here: \url{https://github.com/facebookresearch/ConvNeXt-V2}}.
We set $\lambda$ to 0.05 for computing the least-square weights, $\mathbf{W}_{LS}$ (Eq.~\ref{eq:w_ls}) since ConvNeXtV2-F is pre-trained on ImageNet-1K using the weight decay of 0.05.
We run all experiments on the same hardware with a single GPU (NVIDIA RTX A5000).

\subsection{Main Experiments : Last-Layer Plastic} 
\label{sec:last_layer_details}

We train last-layer classifier while keeping the remaining layers frozen. 
During CL, we use three different loss functions such as CE, MSE, and Squentropy. 
For both CE and Squentropy, we start with the off-the-shelf CE-pretrianed model during CL. For MSE, we perform loss alignment where we fine-tune last-layer classifier on ImageNet-1K dataset (details in Sec.~\ref{sec:loss_align}) and start with MSE-finetuned model during CL.
For MSE loss scaling, we set $\kappa=15$ and $\beta=30$.  
For each CL task, we train model for $1200$ iterations (compute bound). 
During each iteration, model is updated on $256$ samples.
We use AdamW optimizer with weight decay of $0.05$ and fixed learning rate of $0.001$ (we do not use any learning rate scheduler). 
To impose storage constraints, we store maximum 192K samples in the buffer for rehearsal.
We assess performance during rehearsal every 50 iterations to compute the metrics.

\subsection{Main Experiments : Top Layers Plastic}
\label{sec:top_layer_details}

We repeat the training process outlined above for training the last-layer classifier. We change optimizer setting for training top layers including last-layer classifier.
During each training iteration, model is updated on a batch of $256$ samples. We use AdamW optimizer with weight decay of $0.05$ and fixed learning rate of $0.0015$.
The learning rate is reduced in earlier layers by a layer-wise decay factor of $0.9$. We apply OneCycle learning rate scheduler~\citep{smith2017super}.
We control plasticity in top layers using LoRA. Details are given in Sec.~\ref{sec:lora}. For all experiments, we set the rank of the LoRA weight matrices to 48.

\subsection{Controlling Plasticity using LoRA}
\label{sec:lora}

Following~\citet{harun2024overcoming}, we control plasticity in top layers using LoRA. Here we describe how we modify linear layers of a DNN to incorporate LoRA weights and learn them during CL.
For task \( j \), let \( \mathbf{W}_{j-1} \in \mathbb{R}^{d \times g} \) be a previously learned linear layer. At the start of each CL task, we reparameterize this layer by replacing \( \mathbf{W}_{j-1} \) with  
\[
\boldsymbol{\Theta}_j = \mathbf{W}_{j-1} + \mathbf{BA},
\]
where \( \mathbf{B} \in \mathbb{R}^{d \times r} \) and \( \mathbf{A} \in \mathbb{R}^{r \times g} \) are the LoRA adapter parameters with rank \( r \ll \min(d, g) \). 
To initialize LoRA weights, we adhere to LoRA paper~\citep{hu2022lora}. 
Only \( \mathbf{B} \) and \( \mathbf{A} \) are plastic, with \( \mathbf{A} \) initialized with random Gaussian values and \( \mathbf{B} \) initialized to a zero matrix, so \( \mathbf{BA} = \mathbf{0} \) at the beginning of the CL task. At the end of the CL task, the LoRA parameters are folded into the DNN, i.e.,  
\[
\mathbf{W}_j \leftarrow \boldsymbol{\Theta}_j.
\]  
In LoRA experiments, only the last-layer classifier and the LoRA parameters are plastic and updated using rehearsal.

\subsection{Loss Alignment}
\label{sec:loss_align}

We apply random resized cropping and horizontal flipping for data augmentation. Optimization follows an AdamW setup with a cosine decay scheduler. We use learning rate of $0.001$, weight decay of $0.05$, batch size of 512, and training epochs of 50. For MSE loss scaling, we set $\kappa=15$ and $\beta=30$.

\subsection{Linear Probing for Forward Transfer}
\label{sec:linear_probing}

We perform linear probing to measure forward transfer i.e., how learning current CL task helps improve future CL task.
After learning each CL task, we take the learned backbone and attach a linear probe (i.e., a single MLP layer) in the last-layer with number of classes set to 73 (i.e., number of classes in the next CL task). We only train the linear probe while keeping the backbone frozen.
We use AdamW optimizer with a fixed learning rate of $0.001$ and weight decay of $0$. We do not use any learning rate scheduler. In all cases, we use CE loss for linear probing. The probe is trained for $600$ iterations with a batch size of $128$ per iteration. We do not apply any augmentations. The probe is trained solely on on data associated with next CL task (73 places classes). We report best top-1 accuracy (\%).

\subsection{Efficiency Gain}
\label{sec:eff_gain}

We measure \textbf{efficiency gain}, $\mathcal{G}$ which is average efficiency gain or speedup in learning new tasks. 
For $j$'th new task, we measure number of iterations ($I_{d}$) used by data-driven initialization to reach 95\% of the best accuracy obtained by random initialization and then compute efficiency gain by comparing with the number of iterations ($I_{r}$) used by random initialization i.e., $\mathcal{G}_j = I_{r} / I_{d}$. Finally, we compute average gain as $\mathcal{G} = \frac{1}{J} \sum_{j=1}^{J} \mathcal{G}_j$.

\subsection{Ablation Experiments}
\label{sec:ablation_exp}

\textbf{ResNet Architecture.} 
In ResNet18 experiments, we follow \textit{last-layer plastic} setting from Sec.~\ref{sec:last_layer_details}. We use Pytorch's ResNet18 model which is pre-trained on ImageNet-1K (top-1 accuracy 69.75\%). We set $\lambda$ to $10^{-4}$ for computing the least-square weights, $\mathbf{W}_{LS}$ (Eq.~\ref{eq:w_ls}) since ResNet18 is pre-trained on ImageNet-1K using the weight decay of $10^{-4}$.
In CL experiments, we train ResNet18 on 5 disjoint tasks from Places365-Std dataset (73 classes per task) using CE loss. We use AdamW optimizer with weight decay of 0.05 and fixed learning rate of $10^{-4}$. Other details adhere to Sec.~\ref{sec:last_layer_details}.

\textbf{Various CL Methods.}
We explore various CL algorithms to study different weight initialization methods. We study parameter regularization method EWC and rehearsal and knowledge-distillation method DER++. The experimental details are given below.

\textbf{Elastic Weight Consolidation (EWC).} 
In EWC experiments, we train top layers of ConvNeXtV2-F following the procedure outlined in Sec.~\ref{sec:top_layer_details} except not using LoRA. We do not apply rehearsal and train the model entirely on new data.
We use learning rate of $8\times 10^{-4}$ and weight decay of 0.05 for a batch size of 128 new samples. We set $\lambda$ (regularization coefficient in equation 3 in the original paper) to $10^{6}$ for EWC loss. Other implementation details adhere to EWC paper~\citep{kirkpatrick2017overcoming}.

\textbf{Dark Experience Replay++ (DER++).}
In DER++ experiments, we train top layers of ConvNeXtV2-F following the procedure outlined in Sec.~\ref{sec:top_layer_details}. We control plasticity using LoRA. We set learning rate to $1.5\times 10^{-3}$ and weight decay to 0.05 for a batch size of 256 samples.
We employ $\alpha=0.1$ and $\beta=0.9$ (regularization coefficients) for DER++ loss (Equation 6 in the original paper). Additional implementation details follow DER++ paper~\citep{buzzega2020dark}.

\section{Neural Collapse Metrics}
\label{sec:nc_metrics}

Neural Collapse (NC) describes a structured organization of representations in DNNs~\citep{papyan2020prevalence, kothapalli2023neural, zhu2021geometric, rangamani2023feature}.
The following four criteria characterize Neural Collapse:
\begin{enumerate}
    \item \textbf{Feature Collapse} ($\mathcal{NC}1$): Features within each class concentrate around a single mean, with almost no variability within classes.
    \item \textbf{Simplex ETF Structure} ($\mathcal{NC}2$): Class means, when centered at the global mean, are linearly separable, maximally distant, and form a symmetrical structure on a hypersphere known as a Simplex Equiangular Tight Frame (Simplex ETF).
    \item \textbf{Self-Duality} ($\mathcal{NC}3$): The last-layer classifiers align closely with their corresponding class means, forming a self-dual configuration.
    \item \textbf{Nearest Class Mean Decision} ($\mathcal{NC}4$): The classifier operates similarly to the nearest class-center (NCC) decision rule, assigning classes based on proximity to the class means. 
\end{enumerate}

Here, we describe each NC metric used in our results. Let \( \boldsymbol{\mu}_G \) denote the global mean and \( \boldsymbol{\mu}_c \) the \( c \)-th class mean of the features, \( \{\mathbf{z}_{c,i}\} \) at layer \( l \), defined as follows:
\[
\boldsymbol{\mu}_G = \frac{1}{nC} \sum_{c=1}^C \sum_{i=1}^n \mathbf{z}_{c,i}, \quad \boldsymbol{\mu}_c = \frac{1}{n} \sum_{i=1}^n \mathbf{z}_{c,i} \quad (1 \leq c \leq C).
\]
We drop the layer index \( l \) from notation for simplicity.
Also bias is excluded for notation simplicity. Feature dimension is $d$ instead of $d+1$.

\noindent
\textbf{Within-Class Variability Collapse ($\mathcal{NC}1$):}  
It measures the relative size of the within-class covariance \( \boldsymbol{\Sigma}_W \) with respect to the between-class covariance \( \boldsymbol{\Sigma}_B \) of the DNN features:
\[
\boldsymbol{\Sigma}_W = \frac{1}{nC} \sum_{c=1}^C \sum_{i=1}^n \left( \mathbf{z}_{c,i} - \boldsymbol{\mu}_c \right) \left( \mathbf{z}_{c,i} - \boldsymbol{\mu}_c \right)^\top \in \mathbb{R}^{d \times d},
\]
\[
\boldsymbol{\Sigma}_B = \frac{1}{C} \sum_{c=1}^C \left( \boldsymbol{\mu}_c - \boldsymbol{\mu}_G \right) \left( \boldsymbol{\mu}_c - \boldsymbol{\mu}_G \right)^\top \in \mathbb{R}^{d \times d}.
\]

The $\mathcal{NC}1$ metric is defined as:

\[
\mathcal{NC}1 = \frac{1}{C} \operatorname{trace} \left( \boldsymbol{\Sigma_W} \boldsymbol{\Sigma}_B^{\dagger} \right),
\]
where \( \boldsymbol{\Sigma}_B^{\dagger} \) is the pseudo-inverse of \( \boldsymbol{\Sigma}_B \). Note that $\mathcal{NC}1$ is the most dominant indicator of neural collapse.

\noindent
\textbf{Convergence to Simplex ETF ($\mathcal{NC}2$):}  
It quantifies the \( \ell_2 \) distance between the normalized simplex ETF and the normalized \( \mathbf{WW}^\top \), as follows:
\[
\mathcal{NC}2 := \left\| \frac{\mathbf{WW}^\top}{\| \mathbf{WW}^\top \|_F} - \frac{1}{\sqrt{C-1}} \left( \mathbf{I}_C - \frac{1}{C} \mathbf{1}_C \mathbf{1}_C^\top \right) \right\|_F,
\]
where \( \mathbf{W} \in \mathbb{R}^{C \times d} \) is the weight matrix of the learned classifier.

\noindent
\textbf{Convergence to Self-Duality ($\mathcal{NC}3$):}  
It measures the \( \ell_2 \) distance between the normalized simplex ETF and the normalized \( \mathbf{WZ} \):
\[
\mathcal{NC}3 := \left\| \frac{\mathbf{WZ}}{\| \mathbf{WZ} \|_F} - \frac{1}{\sqrt{C-1}} \left( \mathbf{I}_C - \frac{1}{C} \mathbf{1}_C \mathbf{1}_C^\top \right) \right\|_F,
\]
where \( \mathbf{Z} = \left[ \mathbf{z}_1 - \boldsymbol{\mu}_G \; \cdots \; \mathbf{z}_C - \boldsymbol{\mu}_G \right] \in \mathbb{R}^{d \times C} \) is the centered class-mean matrix.

\textbf{Simplification to NCC ($\mathcal{NC}4$):} It measures the collapse of bias \( \mathbf{b} \):
\[
\mathcal{NC}4 := \left\| \mathbf{b} + \mathbf{W} \boldsymbol{\mu}_G   \right\|_2.
\]


\section{Additional Results}
\label{sec:more_results}

\textbf{Forward Transfer.}
Forward transfer results are given in Table~\ref{tab:forward_transfer}. We observe that data-driven initialization improves forward transfer over random initialization.

\textbf{Qualitative Analysis of New Task Loss \& Accuracy.}
For different weight initializations and loss functions, we illustrate the loss and accuracy dynamics in Figures~\ref{fig:fc_mse},~\ref{fig:fc_sqen},~\ref{fig:sgm_ce},~\ref{fig:sgm_sqen}, and~\ref{fig:top_mse}.
In all cases, data-driven initialization demonstrates efficacy in reducing training loss spikes and improving new task adaptation.

\begin{table}[ht]
\centering
  \caption{\textbf{Comprehensive Results on Forward Transfer}. After training top layers in each CL task, we perform linear probing on the next CL task (73 Places classes) to measure forward transfer.
  }
  \label{tab:forward_transfer}
  \centering
     \begin{tabular}{cccc|cccc|c}
     \hline 
     \multicolumn{1}{c}{Weight Init} &
     \multicolumn{1}{c}{Loss} &
     \multicolumn{1}{c}{$\mathcal{U}$} &
     \multicolumn{1}{c|}{$\mathcal{S}$} &
     \multicolumn{4}{c|}{Linear Probe Accuracy (\%) $\uparrow$} &
     \multicolumn{1}{c}{Avg. $\uparrow$} \\
     & & & & Task 2 & Task 3 & Task 4 & Task 5 & \\
    \toprule
     \textbf{Least Square} &  & & & 60.66 & 63.03 & 59.70 & 66.18 & 62.39 \\
     Random & MSE & 600 & 24K & 54.85 & 56.82 & 54.30 & 60.33 & 56.58 \\
     Class Mean &  & & & 60.47 & 63.44 & 60.56 & 67.04 & \textbf{62.88} \\
     \hline
     \textbf{Least Square} &  & & & 63.73 & 67.05 & 64.37 & 72.05 & \textbf{66.80} \\
     Random & CE & 600 & 24K & 62.07 & 65.71 & 63.45 & 71.07 & 65.58 \\
     Class Mean &  & & & 63.29 & 67.45 & 64.22 & 72.00 & 66.74 \\
     \hline
     \textbf{Least Square} &  & & & 61.93 & 65.33 & 62.73 & 70.11 & \textbf{65.03} \\
     Random & SQEN & 600 & 24K & 60.44 & 63.66 & 61.66 & 69.30 & 63.77 \\
     Class Mean &  & & & 61.49 & 65.55 & 62.52 & 69.78 & 64.84 \\
     \toprule
     \textbf{Least Square} &  &  &  & 61.60 & 64.03 & 60.81 & 67.34 & 63.45 \\
     Random & MSE & 1200 & 192K & 55.66 & 59.33 & 56.52 & 63.18 & 58.67 \\
     Class Mean &  &  &  & 62.48 & 65.26 & 62.16 & 68.53 & \textbf{64.61} \\
     \hline
     \textbf{Least Square} &  &  &  & 64.82 & 68.38 & 65.66 & 73.29 & \textbf{68.04} \\
     Random & CE & 1200 & 192K & 64.70 & 67.88 & 65.30 & 72.86 & 67.69 \\
     Class Mean &  &  &  & 64.62 & 68.38 & 65.93 & 73.07 & 68.00 \\
     \hline
     \textbf{Least Square} &  &  &  & 63.53 & 67.38 & 64.26 & 71.88 & \textbf{66.76} \\
     Random & SQEN & 1200 & 192K & 62.00 & 65.75 & 63.86 & 71.01 & 65.66 \\
     Class Mean &  &  &  & 63.00 & 67.00 & 65.23 & 71.29 & 66.63 \\
    \bottomrule
    \end{tabular}
\end{table}

\input{figures/fc_mse}

\input{figures/fc_sqen}

\input{figures/sgm_ce}

\input{figures/sgm_sqen}

\input{figures/sgm_mse}

\begin{table}[ht]
\centering
  \caption{\textbf{Computational Cost Comparison}. 
  Comparison of computational cost in terms of wall-clock runtime, measured on a single NVIDIA RTX A5000 GPU. We report: (1) the total runtime for the full end-to-end CL experiment across five tasks, and (2) the runtime spent solely on weight initialization. In this experiment, ConvNeXtV2-F models, pre-trained on ImageNet-1K, incrementally learn five disjoint tasks (73 Places classes per task) in CL with controlled plasticity setting ($\mathcal{U}=1200$ \& $\mathcal{S}=192K$). Time is reported in minutes. 
  } 
  \label{tab:time}
  \centering
     \begin{tabular}{cc|cc|c}
     \toprule
     \multicolumn{1}{c}{\textbf{Weight Init}} &
     \multicolumn{1}{c|}{\textbf{Loss}} &
     \multicolumn{2}{c|}{\textbf{Time (Mins)} $\downarrow$} &
     \multicolumn{1}{c}{\% \textbf{Increase}} \\
     & & Total & Init Only & \\
    \toprule
     Random &  & 186 & -- & -- \\
     Class Mean & CE & 196 & 10 & +0.05\% \\
     \textbf{Least Square} &  & 199 & 13 & +0.07\% \\
     \hline
     Random &  & 193 & -- & -- \\
     Class Mean & SQEN & 203 & 10 & +0.05\% \\
     \textbf{Least Square} &  & 206 & 13 & +0.07\% \\
     \hline
     Random &  & 194 & -- & -- \\
     Class Mean & MSE & 204 & 10 & +0.05\% \\
     \textbf{Least Square} &  & 207 & 13 & +0.07\% \\
    \bottomrule
    \end{tabular}
\end{table}


%% file: figures/fc_mse.tex
\begin{figure}[t]
  \centering

\begin{subfigure}[b]{0.48\textwidth}
         \centering
         \includegraphics[width=\textwidth]{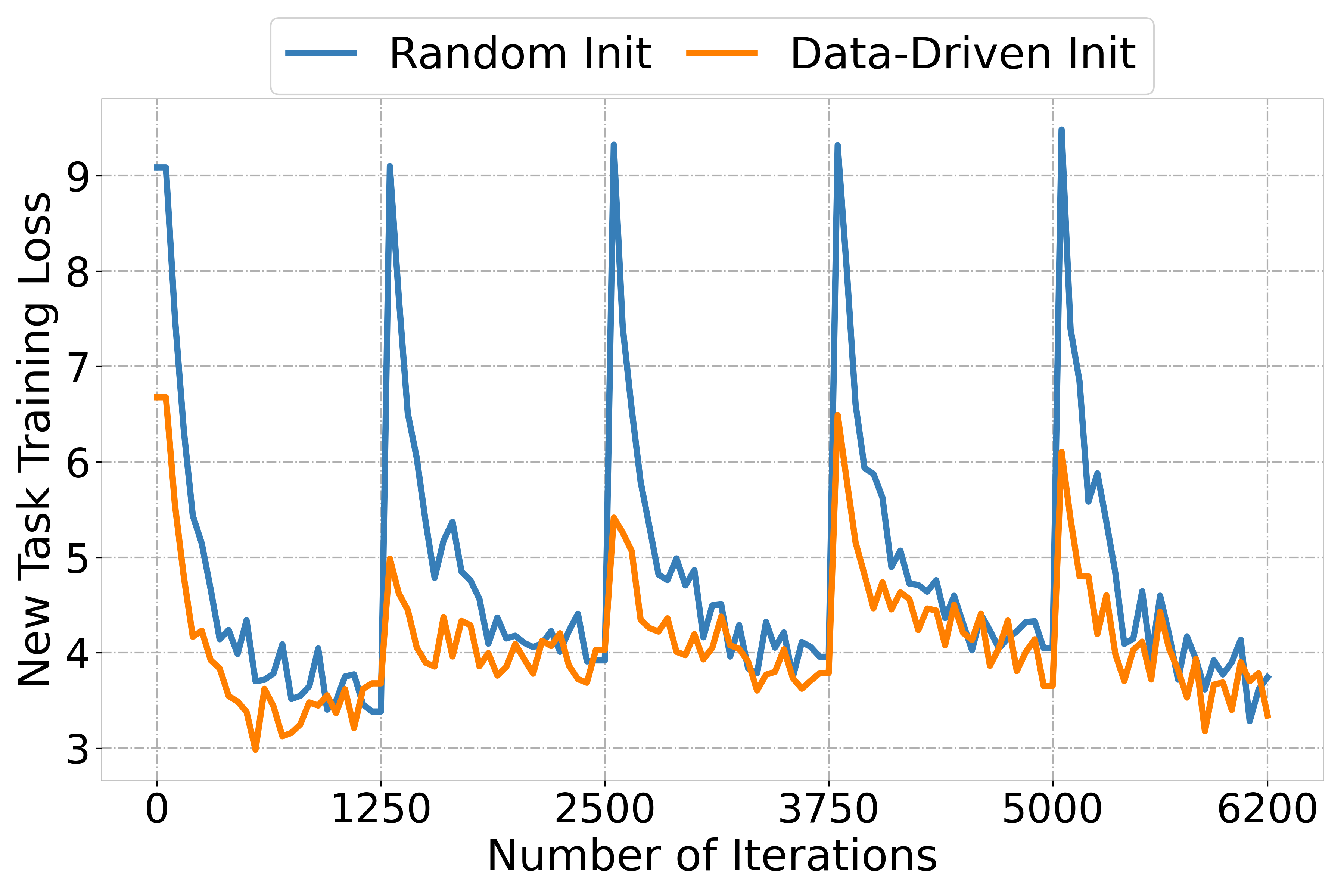}
         \caption{Training Loss}
         \label{fig:train_loss_mse}
     \end{subfigure}
     \hfill
      \begin{subfigure}[b]{0.48\textwidth}
         \centering
         \includegraphics[width=\textwidth]{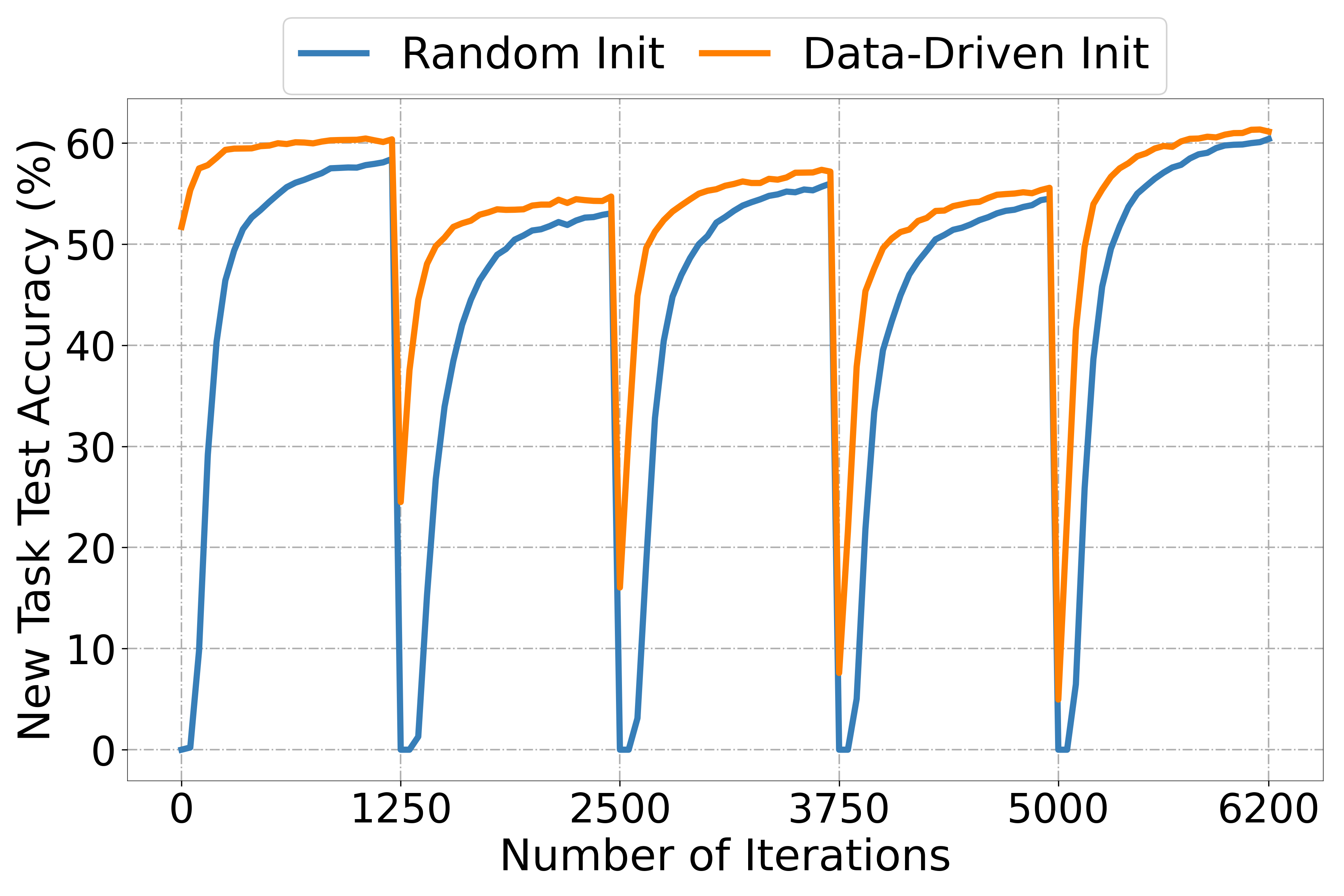}
         \caption{Test Accuracy}
         \label{fig:test_acc_mse}
     \end{subfigure}
   \caption{\textbf{Last-layer trained with MSE}. Data-driven weight initialization (LS) mitigates training loss spikes and improves new task accuracy compared to random initialization. MSE uses scaling parameters.
   }
   \label{fig:fc_mse}
\end{figure}

%% file: figures/fc_sqen.tex
\begin{figure}[t]
  \centering

\begin{subfigure}[b]{0.48\textwidth}
         \centering
         \includegraphics[width=\textwidth]{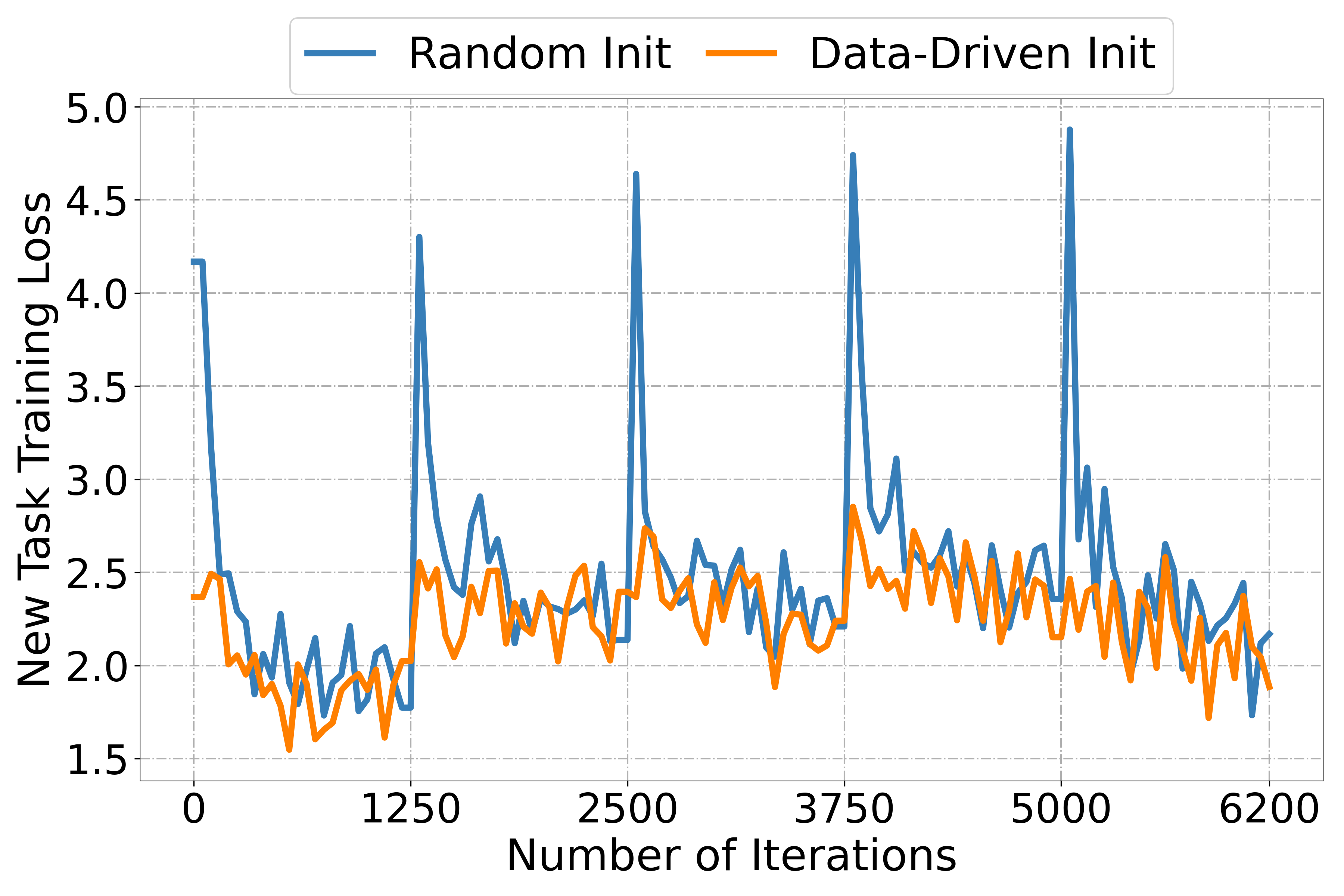}
         \caption{Training Loss}
         \label{fig:train_loss_sqen}
     \end{subfigure}
     \hfill
      \begin{subfigure}[b]{0.48\textwidth}
         \centering
         \includegraphics[width=\textwidth]{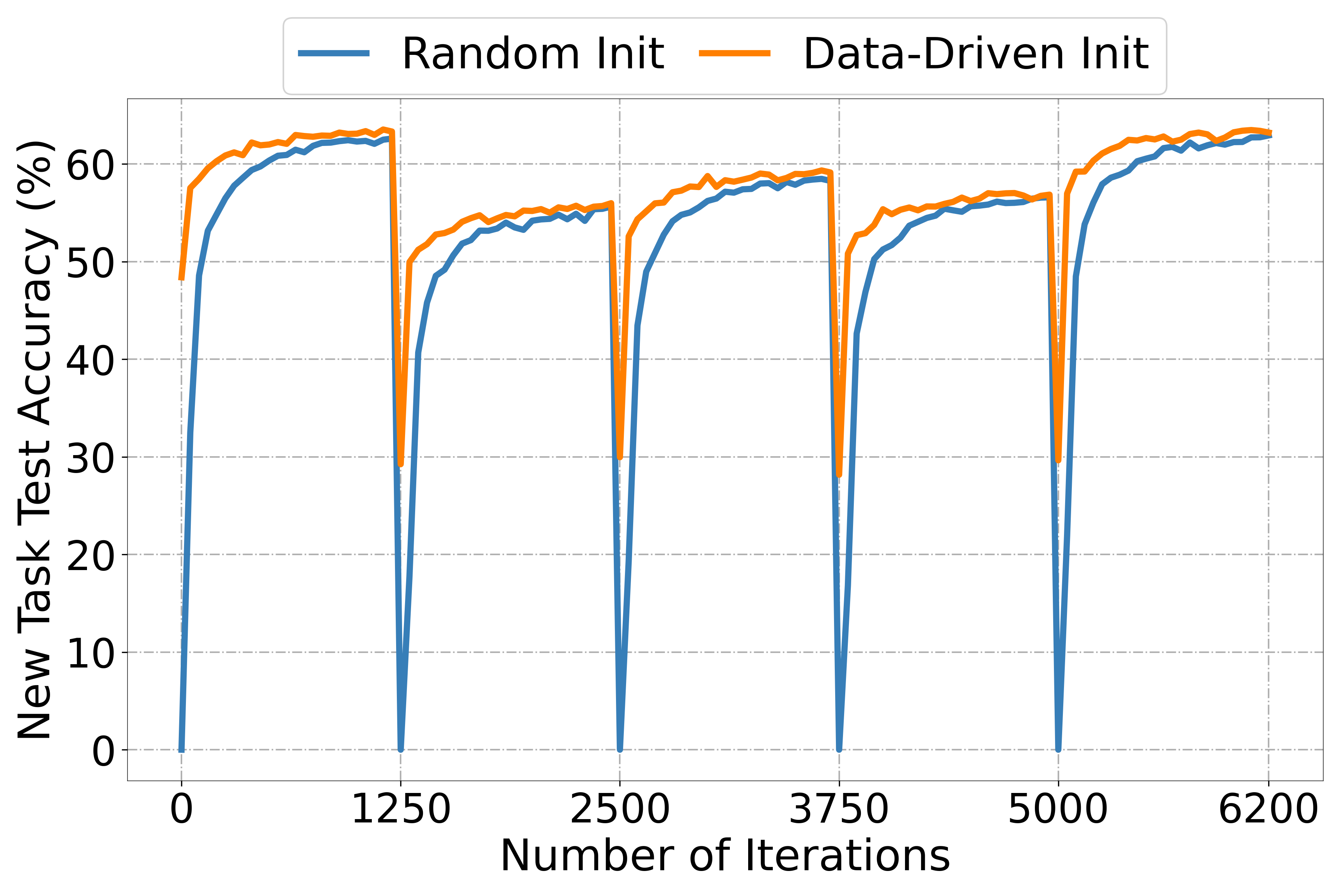}
         \caption{Test Accuracy}
         \label{fig:test_acc_sqen}
     \end{subfigure}
   \caption{\textbf{Last-layer trained with squentropy loss}. Data-driven weight initialization mitigates training loss spikes and improves new task accuracy compared to random initialization.
   }
   \label{fig:fc_sqen}
\end{figure}

%% file: figures/sgm_ce.tex
\begin{figure}[t]
  \centering

\begin{subfigure}[b]{0.48\textwidth}
         \centering
         \includegraphics[width=\textwidth]{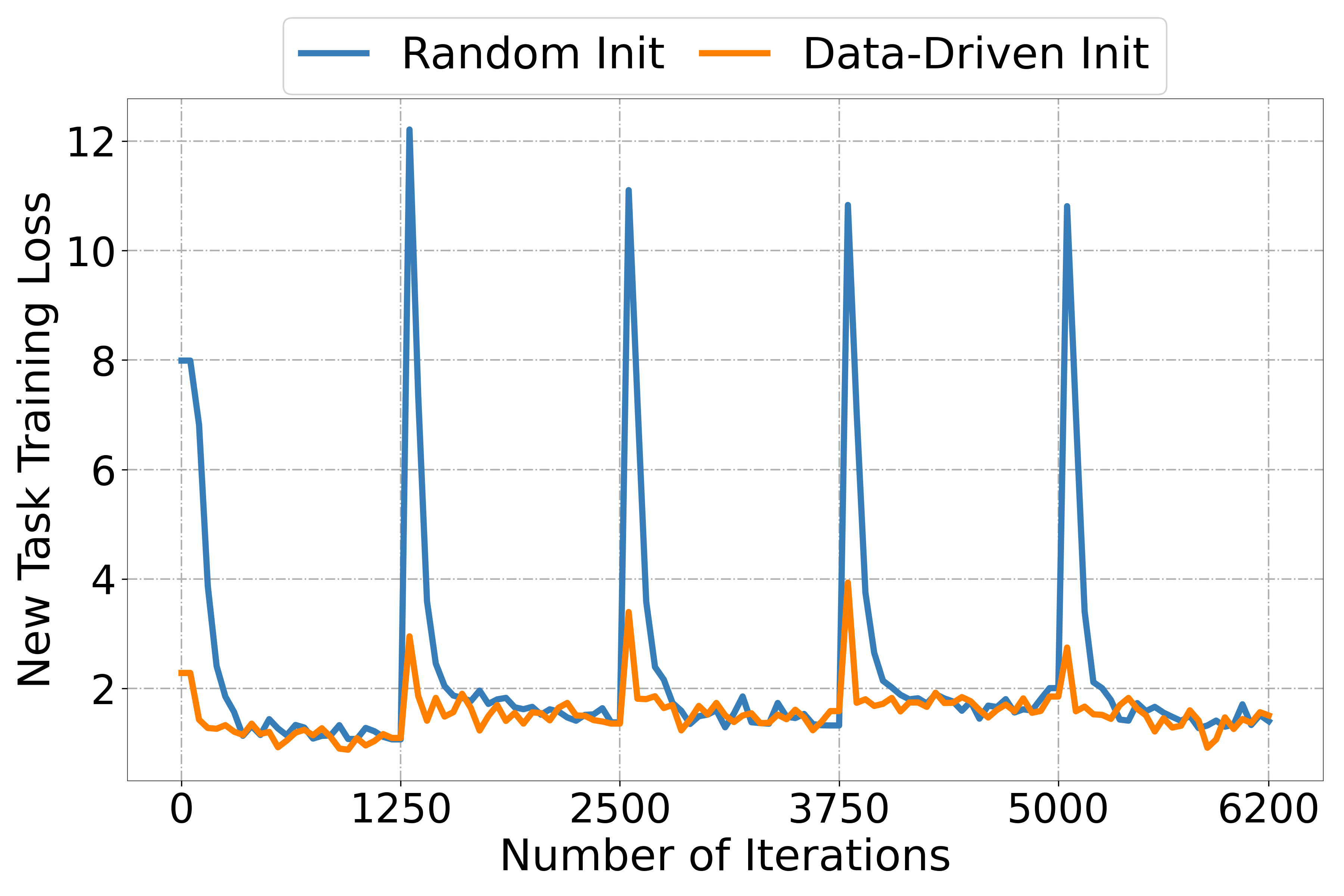}
         \caption{Training Loss}
         \label{fig:train_loss_sgm_ce}
     \end{subfigure}
     \hfill
      \begin{subfigure}[b]{0.48\textwidth}
         \centering
         \includegraphics[width=\textwidth]{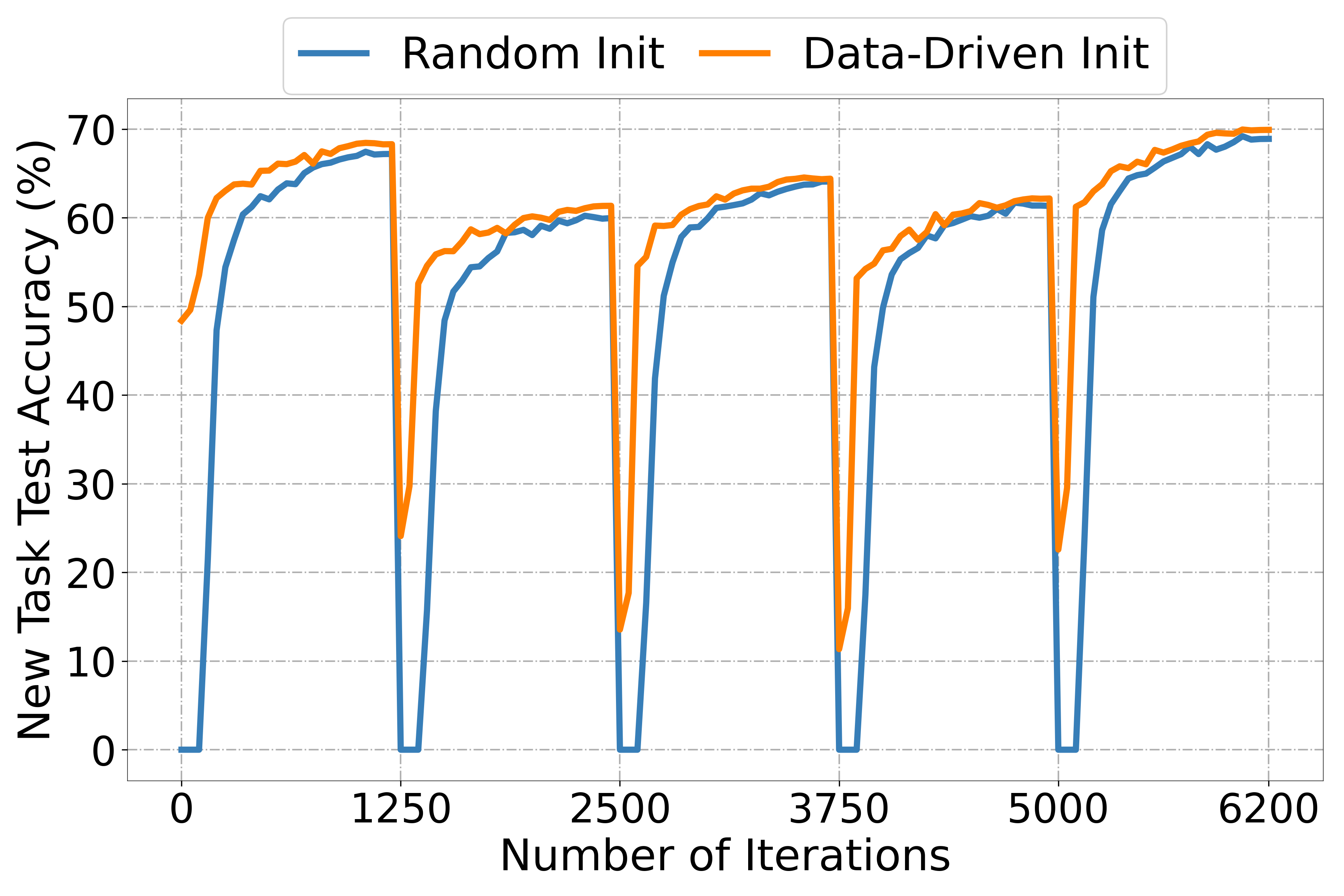}
         \caption{Test Accuracy}
         \label{fig:test_acc_sgm_ce}
     \end{subfigure}
   \caption{\textbf{Top layers trained with cross-entropy loss}. Data-driven weight initialization mitigates training loss spikes and improves new task accuracy compared to random initialization.
   }
   \label{fig:sgm_ce}
\end{figure}

%% file: figures/sgm_sqen.tex
\begin{figure}[t]
  \centering

\begin{subfigure}[b]{0.48\textwidth}
         \centering
         \includegraphics[width=\textwidth]{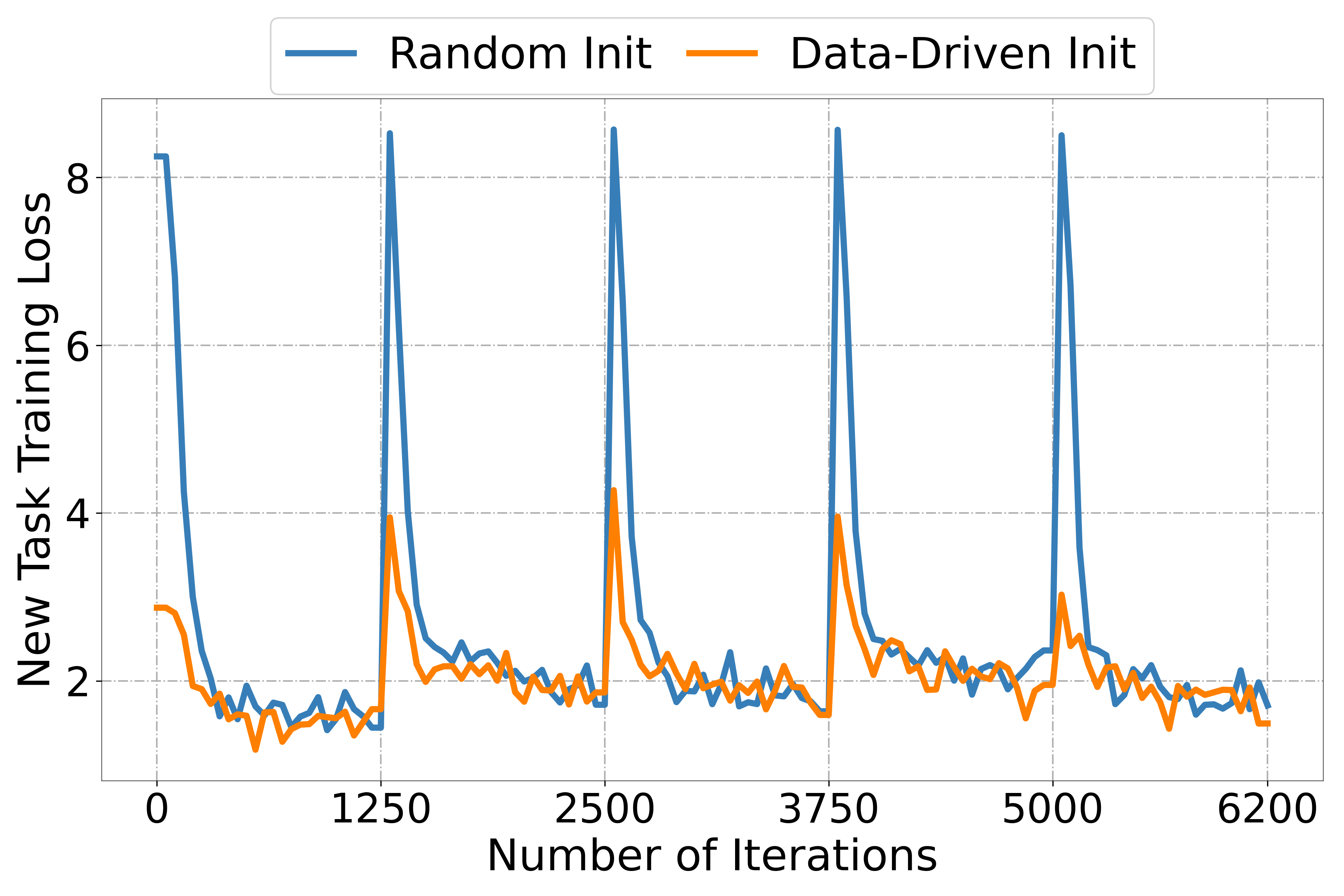}
         \caption{Training Loss}
         \label{fig:train_loss_sgm_sqen}
     \end{subfigure}
     \hfill
      \begin{subfigure}[b]{0.48\textwidth}
         \centering
         \includegraphics[width=\textwidth]{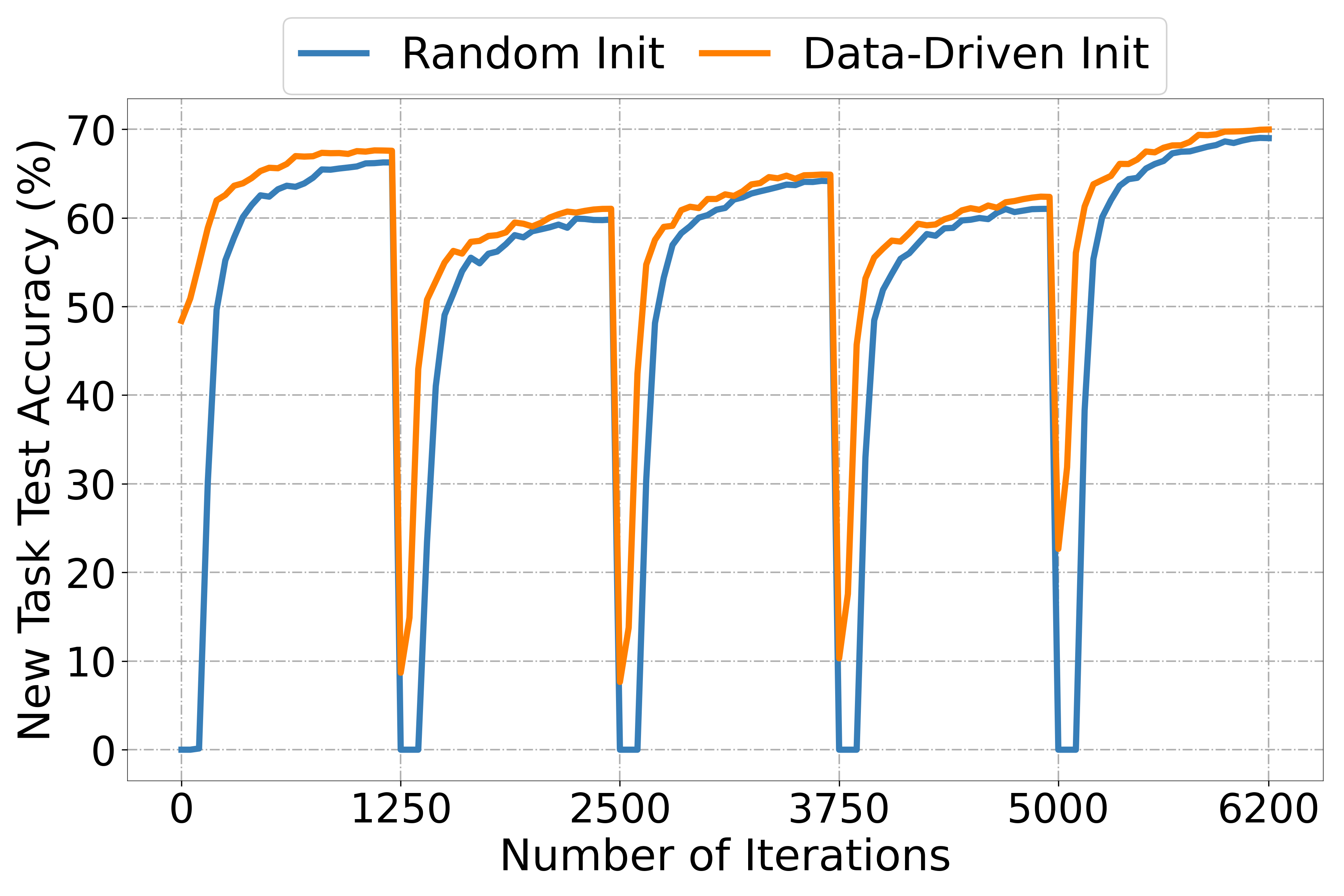}
         \caption{Test Accuracy}
         \label{fig:test_acc_sgm_sqen}
     \end{subfigure}
   \caption{\textbf{Top layers trained with squentropy loss}. Data-driven weight initialization mitigates training loss spikes and improves new task accuracy compared to random initialization.
   }
   \label{fig:sgm_sqen}
\end{figure}

%% file: figures/sgm_mse.tex
\begin{figure}[t]
  \centering

\begin{subfigure}[b]{0.48\textwidth}
         \centering
         \includegraphics[width=\textwidth]{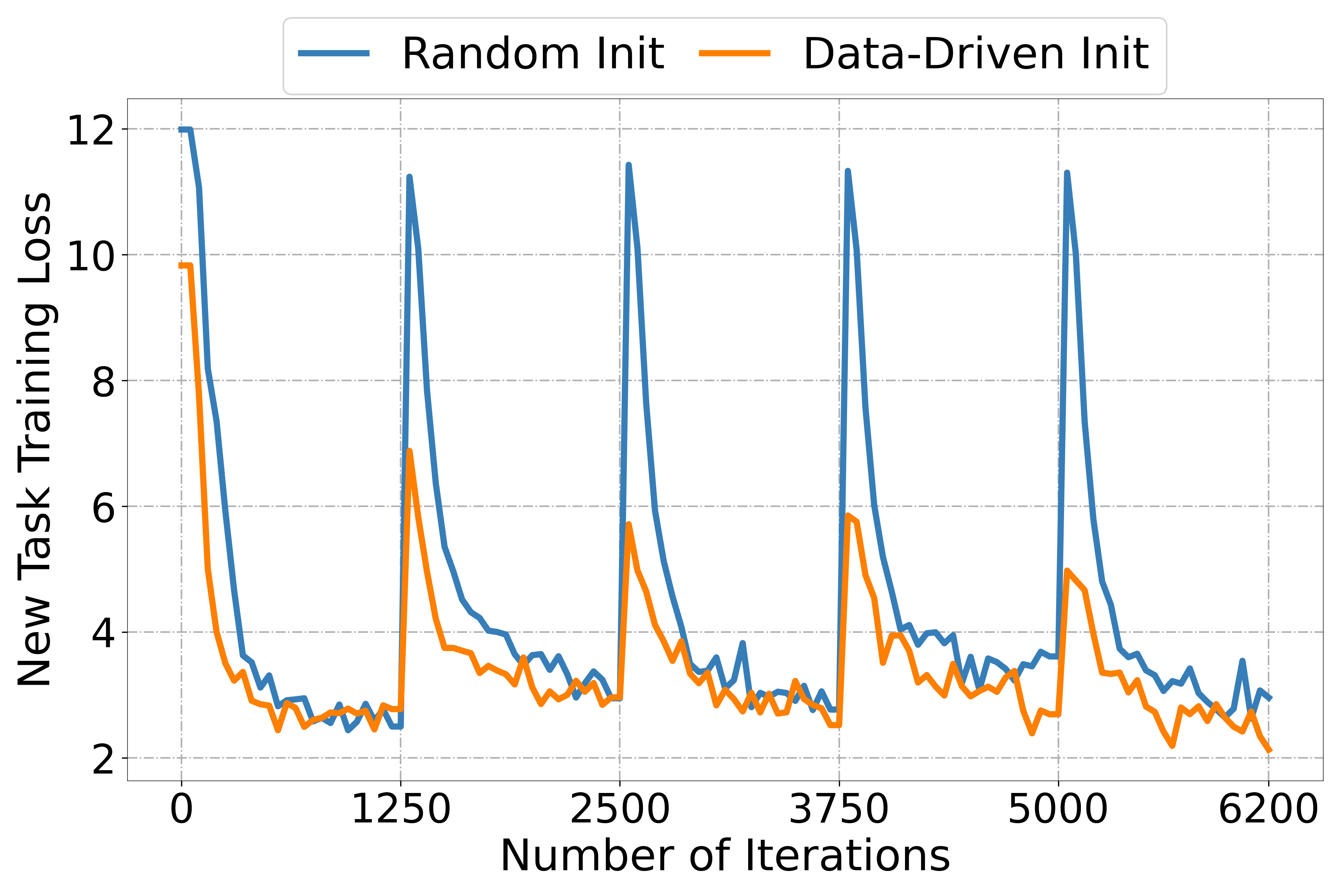}
         \caption{Training Loss}
         \label{fig:train_loss_top_mse}
     \end{subfigure}
     \hfill
      \begin{subfigure}[b]{0.48\textwidth}
         \centering
         \includegraphics[width=\textwidth]{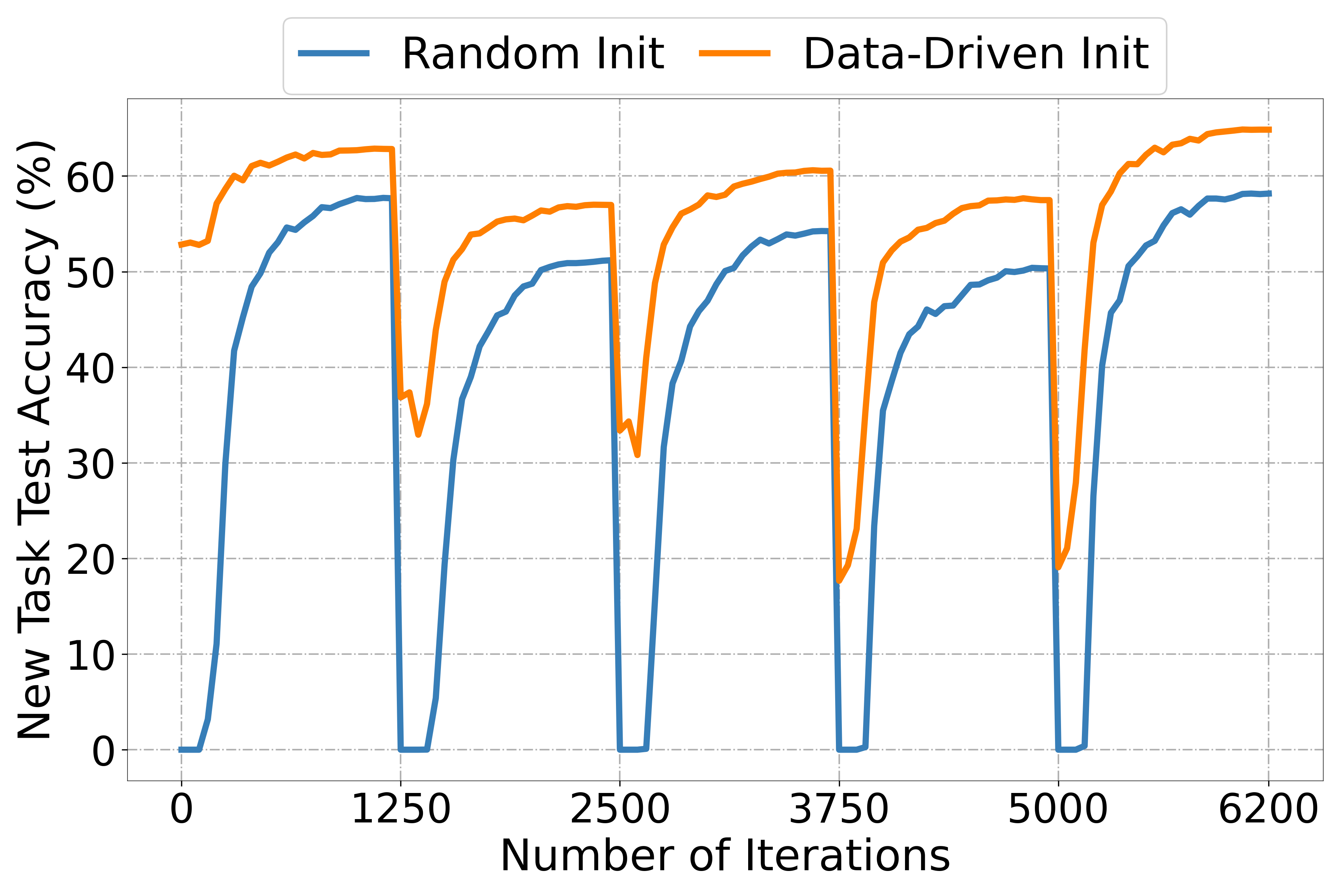}
         \caption{Test Accuracy}
         \label{fig:test_acc_top_mse}
     \end{subfigure}
   \caption{\textbf{Top layers trained with MSE}. Data-driven weight initialization (least-square) mitigates training loss spikes and improves new task accuracy compared to random initialization. MSE uses scaling parameters.
   }
   \label{fig:top_mse}
\end{figure}

%% file: collas2025_conference.bbl
\begin{thebibliography}{73}
\providecommand{\natexlab}[1]{#1}
\providecommand{\url}[1]{\texttt{#1}}
\expandafter\ifx\csname urlstyle\endcsname\relax
  \providecommand{\doi}[1]{doi: #1}\else
  \providecommand{\doi}{doi: \begingroup \urlstyle{rm}\Url}\fi

\bibitem[Acharya et~al.(2020)Acharya, Hayes, and Kanan]{acharya2020rodeo}
Manoj Acharya, Tyler~L Hayes, and Christopher Kanan.
\newblock Rodeo: Replay for online object detection.
\newblock In \emph{BMVC}, 2020.

\bibitem[Aljundi et~al.(2018)Aljundi, Babiloni, Elhoseiny, Rohrbach, and Tuytelaars]{aljundi2018memory}
Rahaf Aljundi, Francesca Babiloni, Mohamed Elhoseiny, Marcus Rohrbach, and Tinne Tuytelaars.
\newblock Memory aware synapses: Learning what (not) to forget.
\newblock In \emph{Proceedings of the European Conference on Computer Vision (ECCV)}, pp.\  139--154, 2018.

\bibitem[Bachlechner et~al.(2021)Bachlechner, Majumder, Mao, Cottrell, and McAuley]{bachlechner2021rezero}
Thomas Bachlechner, Bodhisattwa~Prasad Majumder, Henry Mao, Gary Cottrell, and Julian McAuley.
\newblock Rezero is all you need: Fast convergence at large depth.
\newblock In \emph{Uncertainty in Artificial Intelligence}, pp.\  1352--1361. PMLR, 2021.

\bibitem[Buzzega et~al.(2020)Buzzega, Boschini, Porrello, Abati, and Calderara]{buzzega2020dark}
Pietro Buzzega, Matteo Boschini, Angelo Porrello, Davide Abati, and Simone Calderara.
\newblock Dark experience for general continual learning: a strong, simple baseline.
\newblock \emph{Advances in neural information processing systems}, 33:\penalty0 15920--15930, 2020.

\bibitem[Chaudhry et~al.(2018)Chaudhry, Dokania, Ajanthan, and Torr]{chaudhry2018riemannian}
Arslan Chaudhry, Puneet~K Dokania, Thalaiyasingam Ajanthan, and Philip~HS Torr.
\newblock Riemannian walk for incremental learning: Understanding forgetting and intransigence.
\newblock In \emph{Proceedings of the European Conference on Computer Vision (ECCV)}, pp.\  532--547, 2018.

\bibitem[Chaudhry et~al.(2019)Chaudhry, Rohrbach, Elhoseiny, Ajanthan, Dokania, Torr, and Ranzato]{chaudhryER_2019}
Arslan Chaudhry, Marcus Rohrbach, Mohamed Elhoseiny, Thalaiyasingam Ajanthan, Puneet~K Dokania, Philip~HS Torr, and Marc’Aurelio Ranzato.
\newblock Continual learning with tiny episodic memories.
\newblock \emph{arXiv preprint arXiv:1902.10486, 2019}, 2019.

\bibitem[De \& Smith(2020)De and Smith]{de2020batch}
Soham De and Sam Smith.
\newblock Batch normalization biases residual blocks towards the identity function in deep networks.
\newblock \emph{Advances in Neural Information Processing Systems}, 33:\penalty0 19964--19975, 2020.

\bibitem[Dhar et~al.(2019)Dhar, Singh, Peng, Wu, and Chellappa]{dhar2019learning}
Prithviraj Dhar, Rajat~Vikram Singh, Kuan-Chuan Peng, Ziyan Wu, and Rama Chellappa.
\newblock Learning without memorizing.
\newblock In \emph{Proceedings of the IEEE/CVF Conference on Computer Vision and Pattern Recognition}, pp.\  5138--5146, 2019.

\bibitem[Douillard et~al.(2021)Douillard, Ram{\'e}, Couairon, and Cord]{douillard2021dytox}
Arthur Douillard, Alexandre Ram{\'e}, Guillaume Couairon, and Matthieu Cord.
\newblock Dytox: Transformers for continual learning with dynamic token expansion.
\newblock \emph{arXiv preprint arXiv:2111.11326}, 2021.

\bibitem[Gama et~al.(2014)Gama, {\v{Z}}liobait{\.e}, Bifet, Pechenizkiy, and Bouchachia]{gama2014survey}
Jo{\~a}o Gama, Indr{\.e} {\v{Z}}liobait{\.e}, Albert Bifet, Mykola Pechenizkiy, and Abdelhamid Bouchachia.
\newblock A survey on concept drift adaptation.
\newblock \emph{ACM computing surveys (CSUR)}, 46\penalty0 (4):\penalty0 1--37, 2014.

\bibitem[Gao et~al.(2023)Gao, Zhao, Sun, Xi, Zhang, Ghanem, and Zhang]{gao2023lae}
Qiankun Gao, Chen Zhao, Yifan Sun, Teng Xi, Gang Zhang, Bernard Ghanem, and Jian Zhang.
\newblock A unified continual learning framework with general parameter-efficient tuning.
\newblock \emph{International Conference on Computer Vision (ICCV)}, 2023.

\bibitem[Ghunaim et~al.(2023)Ghunaim, Bibi, Alhamoud, Alfarra, Hammoud, Prabhu, Torr, and Ghanem]{ghunaim2023real}
Yasir Ghunaim, Adel Bibi, Kumail Alhamoud, Motasem Alfarra, Hasan Abed Al~Kader Hammoud, Ameya Prabhu, Philip~HS Torr, and Bernard Ghanem.
\newblock Real-time evaluation in online continual learning: A new paradigm.
\newblock In \emph{CVPR}, 2023.

\bibitem[Glorot \& Bengio(2010)Glorot and Bengio]{glorot2010understanding}
Xavier Glorot and Yoshua Bengio.
\newblock Understanding the difficulty of training deep feedforward neural networks.
\newblock In \emph{Proceedings of the thirteenth international conference on artificial intelligence and statistics}, pp.\  249--256. JMLR Workshop and Conference Proceedings, 2010.

\bibitem[Han et~al.(2022)Han, Papyan, and Donoho]{han2022neural}
X.Y. Han, Vardan Papyan, and David~L. Donoho.
\newblock Neural collapse under {MSE} loss: Proximity to and dynamics on the central path.
\newblock In \emph{International Conference on Learning Representations}, 2022.
\newblock URL \url{https://openreview.net/forum?id=w1UbdvWH_R3}.

\bibitem[Hardt \& Ma(2017)Hardt and Ma]{hardt2017identity}
Moritz Hardt and Tengyu Ma.
\newblock Identity matters in deep learning.
\newblock In \emph{International Conference on Learning Representations}, 2017.

\bibitem[Harun \& Kanan(2024)Harun and Kanan]{harun2024overcoming}
Md~Yousuf Harun and Christopher Kanan.
\newblock Overcoming the stability gap in continual learning.
\newblock \emph{Transactions on Machine Learning Research}, 2024.
\newblock ISSN 2835-8856.
\newblock URL \url{https://openreview.net/forum?id=o2wEfwUOma}.

\bibitem[Harun et~al.(2023{\natexlab{a}})Harun, Gallardo, Hayes, and Kanan]{harun2023efficient}
Md~Yousuf Harun, Jhair Gallardo, Tyler~L. Hayes, and Christopher Kanan.
\newblock How efficient are today's continual learning algorithms?
\newblock In \emph{Proceedings of the IEEE/CVF Conference on Computer Vision and Pattern Recognition (CVPR) Workshops}, pp.\  2431--2436, June 2023{\natexlab{a}}.

\bibitem[Harun et~al.(2023{\natexlab{b}})Harun, Gallardo, Hayes, Kemker, and Kanan]{harun2023siesta}
Md~Yousuf Harun, Jhair Gallardo, Tyler~L. Hayes, Ronald Kemker, and Christopher Kanan.
\newblock {SIESTA}: Efficient online continual learning with sleep.
\newblock \emph{Transactions on Machine Learning Research}, 2023{\natexlab{b}}.
\newblock ISSN 2835-8856.
\newblock URL \url{https://openreview.net/forum?id=MqDVlBWRRV}.

\bibitem[Harun et~al.(2024{\natexlab{a}})Harun, Gallardo, Chen, and Kanan]{harun2024grasp}
Md~Yousuf Harun, Jhair Gallardo, Junyu Chen, and Christopher Kanan.
\newblock Grasp: A rehearsal policy for efficient online continual learning.
\newblock In \emph{CoLLAs}, 2024{\natexlab{a}}.

\bibitem[Harun et~al.(2024{\natexlab{b}})Harun, Lee, Gallardo, Krishnan, and Kanan]{harun2024what}
Md~Yousuf Harun, Kyungbok Lee, Gianmarco Gallardo, Giri Krishnan, and Christopher Kanan.
\newblock What variables affect out-of-distribution generalization in pretrained models?
\newblock \emph{Advances in Neural Information Processing Systems}, 37:\penalty0 56479--56525, 2024{\natexlab{b}}.

\bibitem[Harun et~al.(2025)Harun, Gallardo, and Kanan]{harun2025controlling}
Md~Yousuf Harun, Jhair Gallardo, and Christopher Kanan.
\newblock Controlling neural collapse enhances out-of-distribution detection and transfer learning.
\newblock In \emph{International Conference on Machine Learning}, 2025.

\bibitem[Hayes \& Kanan(2022)Hayes and Kanan]{hayes2022online}
Tyler~L Hayes and Christopher Kanan.
\newblock Online continual learning for embedded devices.
\newblock In \emph{CoLLAs}, 2022.

\bibitem[He et~al.(2015)He, Zhang, Ren, and Sun]{he2015delving}
Kaiming He, Xiangyu Zhang, Shaoqing Ren, and Jian Sun.
\newblock Delving deep into rectifiers: Surpassing human-level performance on imagenet classification.
\newblock In \emph{Proceedings of the IEEE international conference on computer vision}, pp.\  1026--1034, 2015.

\bibitem[He et~al.(2016)He, Zhang, Ren, and Sun]{he2016deep}
Kaiming He, Xiangyu Zhang, Shaoqing Ren, and Jian Sun.
\newblock Deep residual learning for image recognition.
\newblock In \emph{Proceedings of the IEEE conference on computer vision and pattern recognition}, pp.\  770--778, 2016.

\bibitem[Hou et~al.(2019)Hou, Pan, Loy, Wang, and Lin]{hou2019learning}
Saihui Hou, Xinyu Pan, Chen~Change Loy, Zilei Wang, and Dahua Lin.
\newblock Learning a unified classifier incrementally via rebalancing.
\newblock In \emph{Proceedings of the IEEE/CVF Conference on Computer Vision and Pattern Recognition}, pp.\  831--839, 2019.

\bibitem[Hu et~al.(2022)Hu, yelong shen, Wallis, Allen-Zhu, Li, Wang, Wang, and Chen]{hu2022lora}
Edward~J Hu, yelong shen, Phillip Wallis, Zeyuan Allen-Zhu, Yuanzhi Li, Shean Wang, Lu~Wang, and Weizhu Chen.
\newblock Lo{RA}: Low-rank adaptation of large language models.
\newblock In \emph{International Conference on Learning Representations}, 2022.
\newblock URL \url{https://openreview.net/forum?id=nZeVKeeFYf9}.

\bibitem[Hui \& Belkin(2021)Hui and Belkin]{hui2021evaluation}
Like Hui and Mikhail Belkin.
\newblock Evaluation of neural architectures trained with square loss vs cross-entropy in classification tasks.
\newblock In \emph{International Conference on Learning Representations}, 2021.
\newblock URL \url{https://openreview.net/forum?id=hsFN92eQEla}.

\bibitem[Hui et~al.(2023)Hui, Belkin, and Wright]{hui2023cut}
Like Hui, Mikhail Belkin, and Stephen Wright.
\newblock Cut your losses with squentropy.
\newblock In \emph{International Conference on Machine Learning}, pp.\  14114--14131. PMLR, 2023.

\bibitem[Jang et~al.(2022)Jang, Ye, Yang, Shin, Han, Gyeonghun, Choi, and Seo]{jangtowards}
Joel Jang, Seonghyeon Ye, Sohee Yang, Joongbo Shin, Janghoon Han, KIM Gyeonghun, Stanley~Jungkyu Choi, and Minjoon Seo.
\newblock Towards continual knowledge learning of language models.
\newblock In \emph{International Conference on Learning Representations}, 2022.

\bibitem[Jha et~al.(2024)Jha, Gong, and Yao]{jha2024clapclip}
Saurav Jha, Dong Gong, and Lina Yao.
\newblock {CLAP}4{CLIP}: Continual learning with probabilistic finetuning for vision-language models.
\newblock In \emph{The Thirty-eighth Annual Conference on Neural Information Processing Systems}, 2024.
\newblock URL \url{https://openreview.net/forum?id=rF1YRtZfoJ}.

\bibitem[Kirkpatrick et~al.(2017)Kirkpatrick, Pascanu, Rabinowitz, Veness, Desjardins, Rusu, Milan, Quan, Ramalho, Grabska-Barwinska, et~al.]{kirkpatrick2017overcoming}
James Kirkpatrick, Razvan Pascanu, Neil Rabinowitz, Joel Veness, Guillaume Desjardins, Andrei~A Rusu, Kieran Milan, John Quan, Tiago Ramalho, Agnieszka Grabska-Barwinska, et~al.
\newblock Overcoming catastrophic forgetting in neural networks.
\newblock \emph{Proceedings of the national academy of sciences}, 114\penalty0 (13):\penalty0 3521--3526, 2017.

\bibitem[Koh et~al.(2022)Koh, Kim, Ha, and Choi]{koh2022online}
Hyunseo Koh, Dahyun Kim, Jung-Woo Ha, and Jonghyun Choi.
\newblock Online continual learning on class incremental blurry task configuration with anytime inference.
\newblock In \emph{International Conference on Learning Representations}, 2022.
\newblock URL \url{https://openreview.net/forum?id=nrGGfMbY_qK}.

\bibitem[Kornblith et~al.(2021)Kornblith, Chen, Lee, and Norouzi]{kornblith2021better}
Simon Kornblith, Ting Chen, Honglak Lee, and Mohammad Norouzi.
\newblock Why do better loss functions lead to less transferable features?
\newblock \emph{Advances in Neural Information Processing Systems}, 34:\penalty0 28648--28662, 2021.

\bibitem[Kothapalli(2023)]{kothapalli2023neural}
Vignesh Kothapalli.
\newblock Neural collapse: A review on modelling principles and generalization.
\newblock \emph{Transactions on Machine Learning Research}, 2023.
\newblock ISSN 2835-8856.
\newblock URL \url{https://openreview.net/forum?id=QTXocpAP9p}.

\bibitem[Le et~al.(2015)Le, Jaitly, and Hinton]{le2015simple}
Quoc~V Le, Navdeep Jaitly, and Geoffrey~E Hinton.
\newblock A simple way to initialize recurrent networks of rectified linear units.
\newblock \emph{arXiv preprint arXiv:1504.00941}, 2015.

\bibitem[Liu et~al.(2019)Liu, Miao, Zhan, Wang, Gong, and Yu]{liu2019large}
Ziwei Liu, Zhongqi Miao, Xiaohang Zhan, Jiayun Wang, Boqing Gong, and Stella~X Yu.
\newblock Large-scale long-tailed recognition in an open world.
\newblock In \emph{Proceedings of the IEEE/CVF conference on computer vision and pattern recognition}, pp.\  2537--2546, 2019.

\bibitem[Lu et~al.(2018)Lu, Liu, Dong, Gu, Gama, and Zhang]{lu2018learning}
Jie Lu, Anjin Liu, Fan Dong, Feng Gu, Joao Gama, and Guangquan Zhang.
\newblock Learning under concept drift: A review.
\newblock \emph{IEEE transactions on knowledge and data engineering}, 31\penalty0 (12):\penalty0 2346--2363, 2018.

\bibitem[Luccioni et~al.(2022)Luccioni, Viguier, and Ligozat]{luccioni2022estimating}
Alexandra~Sasha Luccioni, Sylvain Viguier, and Anne-Laure Ligozat.
\newblock Estimating the carbon footprint of bloom, a 176b parameter language model.
\newblock \emph{arXiv preprint arXiv:2211.02001}, 2022.

\bibitem[Lyle et~al.(2023)Lyle, Zheng, Nikishin, Pires, Pascanu, and Dabney]{lyle2023understanding}
Clare Lyle, Zeyu Zheng, Evgenii Nikishin, Bernardo~Avila Pires, Razvan Pascanu, and Will Dabney.
\newblock Understanding plasticity in neural networks.
\newblock In \emph{International Conference on Machine Learning}, pp.\  23190--23211. PMLR, 2023.

\bibitem[Mallick et~al.(2022)Mallick, Hsieh, Arzani, and Joshi]{mallick2022matchmaker}
Ankur Mallick, Kevin Hsieh, Behnaz Arzani, and Gauri Joshi.
\newblock Matchmaker: Data drift mitigation in machine learning for large-scale systems.
\newblock \emph{Proceedings of Machine Learning and Systems}, 4:\penalty0 77--94, 2022.

\bibitem[McCloskey \& Cohen(1989)McCloskey and Cohen]{mccloskey1989catastrophic}
Michael McCloskey and Neal~J Cohen.
\newblock Catastrophic interference in connectionist networks: The sequential learning problem.
\newblock In \emph{Psychology of learning and motivation}, volume~24, pp.\  109--165. Elsevier, 1989.

\bibitem[McDonnell et~al.(2024)McDonnell, Gong, Parvaneh, Abbasnejad, and van~den Hengel]{mcdonnell2024ranpac}
Mark~D McDonnell, Dong Gong, Amin Parvaneh, Ehsan Abbasnejad, and Anton van~den Hengel.
\newblock Ranpac: Random projections and pre-trained models for continual learning.
\newblock \emph{Advances in Neural Information Processing Systems}, 36, 2024.

\bibitem[Mirzadeh et~al.(2022)Mirzadeh, Chaudhry, Yin, Nguyen, Pascanu, Gorur, and Farajtabar]{mirzadeh2022architecture}
Seyed~Iman Mirzadeh, Arslan Chaudhry, Dong Yin, Timothy Nguyen, Razvan Pascanu, Dilan Gorur, and Mehrdad Farajtabar.
\newblock Architecture matters in continual learning.
\newblock \emph{arXiv preprint arXiv:2202.00275}, 2022.

\bibitem[Pan et~al.(2025)Pan, Wang, Wu, Wang, Zhang, and Xu]{pan2025idinit}
Yu~Pan, Chaozheng Wang, Zekai Wu, Qifan Wang, Min Zhang, and Zenglin Xu.
\newblock {IDI}nit: A universal and stable initialization method for neural network training.
\newblock In \emph{The Thirteenth International Conference on Learning Representations}, 2025.
\newblock URL \url{https://openreview.net/forum?id=LFiaoYnP6T}.

\bibitem[Papyan et~al.(2020)Papyan, Han, and Donoho]{papyan2020prevalence}
Vardan Papyan, XY~Han, and David~L Donoho.
\newblock Prevalence of neural collapse during the terminal phase of deep learning training.
\newblock \emph{Proceedings of the National Academy of Sciences}, 117\penalty0 (40):\penalty0 24652--24663, 2020.

\bibitem[Parisi et~al.(2019)Parisi, Kemker, Part, Kanan, and Wermter]{parisi2019continual}
German~I Parisi, Ronald Kemker, Jose~L Part, Christopher Kanan, and Stefan Wermter.
\newblock Continual lifelong learning with neural networks: A review.
\newblock \emph{Neural Networks}, 113:\penalty0 54--71, 2019.

\bibitem[Patterson et~al.(2021)Patterson, Gonzalez, Le, Liang, Munguia, Rothchild, So, Texier, and Dean]{patterson2021carbon}
David Patterson, Joseph Gonzalez, Quoc Le, Chen Liang, Lluis-Miquel Munguia, Daniel Rothchild, David So, Maud Texier, and Jeff Dean.
\newblock Carbon emissions and large neural network training.
\newblock \emph{arXiv preprint arXiv:2104.10350}, 2021.

\bibitem[Prabhu et~al.(2023)Prabhu, Hammoud, Dokania, Torr, Lim, Ghanem, and Bibi]{prabhu2023computationally}
Ameya Prabhu, Hasan Abed Al~Kader Hammoud, Puneet Dokania, Philip~HS Torr, Ser-Nam Lim, Bernard Ghanem, and Adel Bibi.
\newblock Computationally budgeted continual learning: What does matter?
\newblock In \emph{IEEE/CVF Conference on Computer Vision and Pattern Recognition (CVPR)}, 2023.

\bibitem[Rangamani et~al.(2023)Rangamani, Lindegaard, Galanti, and Poggio]{rangamani2023feature}
Akshay Rangamani, Marius Lindegaard, Tomer Galanti, and Tomaso~A Poggio.
\newblock Feature learning in deep classifiers through intermediate neural collapse.
\newblock In \emph{International Conference on Machine Learning}, pp.\  28729--28745. PMLR, 2023.

\bibitem[Rebuffi et~al.(2017)Rebuffi, Kolesnikov, Sperl, and Lampert]{rebuffi2017icarl}
Sylvestre-Alvise Rebuffi, Alexander Kolesnikov, Georg Sperl, and Christoph~H Lampert.
\newblock icarl: Incremental classifier and representation learning.
\newblock In \emph{Proceedings of the IEEE conference on Computer Vision and Pattern Recognition}, pp.\  2001--2010, 2017.

\bibitem[Russakovsky et~al.(2015)Russakovsky, Deng, Su, Krause, Satheesh, Ma, Huang, Karpathy, Khosla, Bernstein, et~al.]{russakovsky2015imagenet}
Olga Russakovsky, Jia Deng, Hao Su, Jonathan Krause, Sanjeev Satheesh, Sean Ma, Zhiheng Huang, Andrej Karpathy, Aditya Khosla, Michael Bernstein, et~al.
\newblock Imagenet large scale visual recognition challenge.
\newblock \emph{International journal of computer vision}, 115\penalty0 (3):\penalty0 211--252, 2015.

\bibitem[Schwartz et~al.(2020)Schwartz, Dodge, Smith, and Etzioni]{schwartz2020green}
Roy Schwartz, Jesse Dodge, Noah~A Smith, and Oren Etzioni.
\newblock Green {AI}.
\newblock \emph{Communications of the ACM}, 63\penalty0 (12):\penalty0 54--63, 2020.

\bibitem[Smith et~al.(2023)Smith, Karlinsky, Gutta, Cascante-Bonilla, Kim, Arbelle, Panda, Feris, and Kira]{smith2023coda}
James~Seale Smith, Leonid Karlinsky, Vyshnavi Gutta, Paola Cascante-Bonilla, Donghyun Kim, Assaf Arbelle, Rameswar Panda, Rogerio Feris, and Zsolt Kira.
\newblock Coda-prompt: Continual decomposed attention-based prompting for rehearsal-free continual learning.
\newblock In \emph{Proceedings of the IEEE/CVF Conference on Computer Vision and Pattern Recognition}, pp.\  11909--11919, 2023.

\bibitem[Smith \& Topin(2017)Smith and Topin]{smith2017super}
Leslie~N Smith and Nicholay Topin.
\newblock Super-convergence: Very fast training of neural networks using large learning rates. arxiv.
\newblock \emph{arXiv preprint arXiv:1708.07120}, 2017.

\bibitem[Tsymbal(2004)]{tsymbal2004problem}
Alexey Tsymbal.
\newblock The problem of concept drift: definitions and related work.
\newblock \emph{Computer Science Department, Trinity College Dublin}, 106\penalty0 (2):\penalty0 58, 2004.

\bibitem[van~de Ven et~al.(2022)van~de Ven, Tuytelaars, and Tolias]{van2022three}
Gido~M van~de Ven, Tinne Tuytelaars, and Andreas~S Tolias.
\newblock Three types of incremental learning.
\newblock \emph{Nature Machine Intelligence}, pp.\  1--13, 2022.

\bibitem[Verwimp et~al.(2024)Verwimp, Ben-David, Bethge, Cossu, Gepperth, Hayes, H{\"u}llermeier, Kanan, Kudithipudi, Lampert, et~al.]{verwimp2023continual}
Eli Verwimp, Shai Ben-David, Matthias Bethge, Andrea Cossu, Alexander Gepperth, Tyler~L Hayes, Eyke H{\"u}llermeier, Christopher Kanan, Dhireesha Kudithipudi, Christoph~H Lampert, et~al.
\newblock Continual learning: Applications and the road forward.
\newblock \emph{TMLR}, 2024.

\bibitem[Wang et~al.(2022{\natexlab{a}})Wang, Zhang, Ebrahimi, Sun, Zhang, Lee, Ren, Su, Perot, Dy, et~al.]{wang2022dualprompt}
Zifeng Wang, Zizhao Zhang, Sayna Ebrahimi, Ruoxi Sun, Han Zhang, Chen-Yu Lee, Xiaoqi Ren, Guolong Su, Vincent Perot, Jennifer Dy, et~al.
\newblock Dualprompt: Complementary prompting for rehearsal-free continual learning.
\newblock In \emph{European Conference on Computer Vision}, pp.\  631--648. Springer, 2022{\natexlab{a}}.

\bibitem[Wang et~al.(2022{\natexlab{b}})Wang, Zhang, Lee, Zhang, Sun, Ren, Su, Perot, Dy, and Pfister]{wang2022learning}
Zifeng Wang, Zizhao Zhang, Chen-Yu Lee, Han Zhang, Ruoxi Sun, Xiaoqi Ren, Guolong Su, Vincent Perot, Jennifer Dy, and Tomas Pfister.
\newblock Learning to prompt for continual learning.
\newblock In \emph{Proceedings of the IEEE/CVF Conference on Computer Vision and Pattern Recognition}, pp.\  139--149, 2022{\natexlab{b}}.

\bibitem[Woo et~al.(2023)Woo, Debnath, Hu, Chen, Liu, Kweon, and Xie]{woo2023convnext}
Sanghyun Woo, Shoubhik Debnath, Ronghang Hu, Xinlei Chen, Zhuang Liu, In~So Kweon, and Saining Xie.
\newblock Convnext v2: Co-designing and scaling convnets with masked autoencoders.
\newblock \emph{arXiv preprint arXiv:2301.00808}, 2023.

\bibitem[Wu et~al.(2022)Wu, Raghavendra, Gupta, Acun, Ardalani, Maeng, Chang, Aga, Huang, Bai, et~al.]{wu2022sustainable}
Carole-Jean Wu, Ramya Raghavendra, Udit Gupta, Bilge Acun, Newsha Ardalani, Kiwan Maeng, Gloria Chang, Fiona Aga, Jinshi Huang, Charles Bai, et~al.
\newblock Sustainable ai: Environmental implications, challenges and opportunities.
\newblock \emph{Proceedings of Machine Learning and Systems}, 4:\penalty0 795--813, 2022.

\bibitem[Wu et~al.(2019)Wu, Chen, Wang, Ye, Liu, Guo, and Fu]{wu2019large}
Yue Wu, Yinpeng Chen, Lijuan Wang, Yuancheng Ye, Zicheng Liu, Yandong Guo, and Yun Fu.
\newblock Large scale incremental learning.
\newblock In \emph{Proceedings of the IEEE/CVF Conference on Computer Vision and Pattern Recognition}, pp.\  374--382, 2019.

\bibitem[Yan et~al.(2021)Yan, Xie, and He]{yan2021dynamically}
Shipeng Yan, Jiangwei Xie, and Xuming He.
\newblock Der: Dynamically expandable representation for class incremental learning.
\newblock In \emph{Proceedings of the IEEE/CVF Conference on Computer Vision and Pattern Recognition}, pp.\  3014--3023, 2021.

\bibitem[Yoon et~al.(2020)Yoon, Kim, Yang, and Hwang]{yoon2020scalable}
Jaehong Yoon, Saehoon Kim, Eunho Yang, and Sung~Ju Hwang.
\newblock Scalable and order-robust continual learning with additive parameter decomposition.
\newblock In \emph{Eighth International Conference on Learning Representations, ICLR 2020}. ICLR, 2020.

\bibitem[Zhang et~al.(2018)Zhang, Dauphin, and Ma]{zhang2018fixup}
Hongyi Zhang, Yann~N Dauphin, and Tengyu Ma.
\newblock Fixup initialization: Residual learning without normalization.
\newblock In \emph{International Conference on Learning Representations}, 2018.

\bibitem[Zhang et~al.(2023)Zhang, Mohamed, Ghanem, Torr, Bibi, and Elhoseiny]{zhang2023continual}
Wenxuan Zhang, Youssef Mohamed, Bernard Ghanem, Philip Torr, Adel Bibi, and Mohamed Elhoseiny.
\newblock Continual learning on a diet: Learning from sparsely labeled streams under constrained computation.
\newblock In \emph{The Twelfth International Conference on Learning Representations}, 2023.

\bibitem[Zhao et~al.(2022)Zhao, Schaefer, and Anandkumar]{zhao2022zero}
Jiawei Zhao, Florian~Tobias Schaefer, and Anima Anandkumar.
\newblock Zero initialization: Initializing neural networks with only zeros and ones.
\newblock \emph{Transactions on Machine Learning Research}, 2022.
\newblock ISSN 2835-8856.
\newblock URL \url{https://openreview.net/forum?id=1AxQpKmiTc}.

\bibitem[Zhou et~al.(2017)Zhou, Lapedriza, Khosla, Oliva, and Torralba]{zhou2017places}
Bolei Zhou, Agata Lapedriza, Aditya Khosla, Aude Oliva, and Antonio Torralba.
\newblock Places: A 10 million image database for scene recognition.
\newblock \emph{IEEE transactions on pattern analysis and machine intelligence}, 40\penalty0 (6):\penalty0 1452--1464, 2017.

\bibitem[Zhou et~al.(2023)Zhou, Wang, Qi, Ye, Zhan, and Liu]{zhou2023deep}
Da-Wei Zhou, Qi-Wei Wang, Zhi-Hong Qi, Han-Jia Ye, De-Chuan Zhan, and Ziwei Liu.
\newblock Deep class-incremental learning: A survey.
\newblock \emph{arXiv preprint arXiv:2302.03648}, 2023.

\bibitem[Zhou et~al.(2022{\natexlab{a}})Zhou, Li, Ding, You, Qu, and Zhu]{zhou2022optimization}
Jinxin Zhou, Xiao Li, Tianyu Ding, Chong You, Qing Qu, and Zhihui Zhu.
\newblock On the optimization landscape of neural collapse under mse loss: Global optimality with unconstrained features.
\newblock In \emph{International Conference on Machine Learning}, pp.\  27179--27202. PMLR, 2022{\natexlab{a}}.

\bibitem[Zhou et~al.(2022{\natexlab{b}})Zhou, You, Li, Liu, Liu, Qu, and Zhu]{zhou2022all}
Jinxin Zhou, Chong You, Xiao Li, Kangning Liu, Sheng Liu, Qing Qu, and Zhihui Zhu.
\newblock Are all losses created equal: A neural collapse perspective.
\newblock \emph{Advances in Neural Information Processing Systems}, 35:\penalty0 31697--31710, 2022{\natexlab{b}}.

\bibitem[Zhu et~al.(2022)Zhu, Chen, Xie, Li, Zhang, Xue, Tian, Chen, et~al.]{zhu2022boosting}
Yao Zhu, YueFeng Chen, Chuanlong Xie, Xiaodan Li, Rong Zhang, Hui Xue, Xiang Tian, Yaowu Chen, et~al.
\newblock Boosting out-of-distribution detection with typical features.
\newblock \emph{Advances in Neural Information Processing Systems}, 35:\penalty0 20758--20769, 2022.

\bibitem[Zhu et~al.(2021)Zhu, Ding, Zhou, Li, You, Sulam, and Qu]{zhu2021geometric}
Zhihui Zhu, Tianyu Ding, Jinxin Zhou, Xiao Li, Chong You, Jeremias Sulam, and Qing Qu.
\newblock A geometric analysis of neural collapse with unconstrained features.
\newblock \emph{Advances in Neural Information Processing Systems}, 34:\penalty0 29820--29834, 2021.

\end{thebibliography}
